\documentclass[10pt,twocolumn,letterpaper]{article}

\usepackage[pagenumbers]{cvpr} 

\usepackage{xspace}

\usepackage{algorithm}
\usepackage{algpseudocode}

\newcommand{\ours}{LiND-JEPA\xspace}

\usepackage{bm}  

\newcommand{\dimref}{d_{\text{lin}}}

\newcommand{\zref}{\bm{z}^{\text{lin}}}
\newcommand{\zrefhat}{\widehat{\bm{z}}^{\text{lin}}}
\newcommand{\zconthat}{\widehat{\bm{z}}^{\text{nlin}}}
\newcommand{\zcont}{\bm{z}^{\text{nlin}}}
\usepackage{mathtools}

\newtheorem{theorem}{Theorem}
\newtheorem{lemma}[theorem]{Lemma}

\newtheorem{corollary}[theorem]{Corollary}

\definecolor{cvprblue}{rgb}{0.21,0.49,0.74}
\usepackage[pagebackref,breaklinks=true,colorlinks=true,allcolors=cvprblue]{hyperref}

\def\confName{CVPR}
\def\confYear{2027}

\title{\ours: Structured Linear-Nonlinear Dynamics for Latent World Models  }

\author{
Hongxu Shen\textsuperscript{1}\quad
Xiaoyu Liu\textsuperscript{1}\quad
Yuyang Jiang\textsuperscript{1}\quad
Yongqi Zhang\textsuperscript{1}\quad\\
Jian Sun\textsuperscript{1}\quad
Huaxin Pei\textsuperscript{2}\quad
Yiqin Yang\textsuperscript{3}\quad
Qinghua Tao\textsuperscript{1}%
\thanks{Correspondence to Qinghua Tao
(\texttt{qinghua.tao@bit.edu.cn}).}\\[3pt]
\textsuperscript{1}Beijing Institute of Technology\\
\textsuperscript{2}DenseAI\\
\textsuperscript{3}Institute of automation, Chinese academy of science\\
}

\begin{document}
\maketitle

\begin{abstract}
Joint-embedding predictive architectures (JEPAs) provide a framework for learning  world models and planning from visual observations in the learned latent space. A key design choice is how to structure latent representations and model their temporal evolution during predictive rollouts. In this work, we introduce \ours, which partitions the latent space into two jointly learned subspaces with \textbf{Li}near and \textbf{N}onlinear \textbf{D}ynamics. An action-conditioned linear map predicts the evolution of one subspace, while a causal Transformer models history-dependent dynamics in the other. The two branches are trained jointly with a prediction loss and subspace-specific regularization to discourage representation collapse and reduce redundancy in the linear coordinates. Experiments on four image-goal control environments show higher average planning success than the evaluated baselines, particularly for distant goals. Ablation studies support the contributions of the predictive structure and subspace-specific regularization. The source code is available at
\url{https://github.com/shenhongxu123/LiND-JEPA}.

\end{abstract}

\section{Introduction}
Latent world models enable agents to predict the consequences of actions and plan from visual observations in a compact latent 
space \cite{ha2018worldmodels,hafner2020dreamer}. Joint-embedding predictive architectures (JEPAs) provide a framework for learning such models by predicting future representations from observed context \cite{lecun2022path}. 
With action-conditioned prediction, a learned world model  can evaluate candidate action sequences with model predictive control by rolling out latent states and comparing them with the representation of a goal image \cite{assran2025vjepa2}. This supports image-goal planning from offline trajectories without  reward annotation.

A key design question is how to organize the latent space according to the temporal dynamics that the predictor is required to capture. The encoder for the latents and the transition mechanism in such latent space for future prediction are closely coupled: the encoder determines what latent features are available for prediction, while the predictives from the transitioning guide the latents to evolve in particular ways that describe the dynamics of the latent world~\cite{zhang2019solar,maes2026lewm,pariente2026bilinear}. Some latent coordinates can admit a simple transition rule~\cite{watter2015embed}, while others may require nonlinear transitions conditioned with recent history {for rich temporal contextuals~\cite{zhou2025dinowm}}. Assigning distinct predictive structures to these coordinates offers a way to organize latent representations and guide their learning across temporal evolution. This is particularly relevant  for planning, where the learned transitions are applied repeatedly to evaluate future action sequences  during multi-step rollouts.

\begin{figure*}[t]
    \centering
    \includegraphics[width=0.95\textwidth]
    {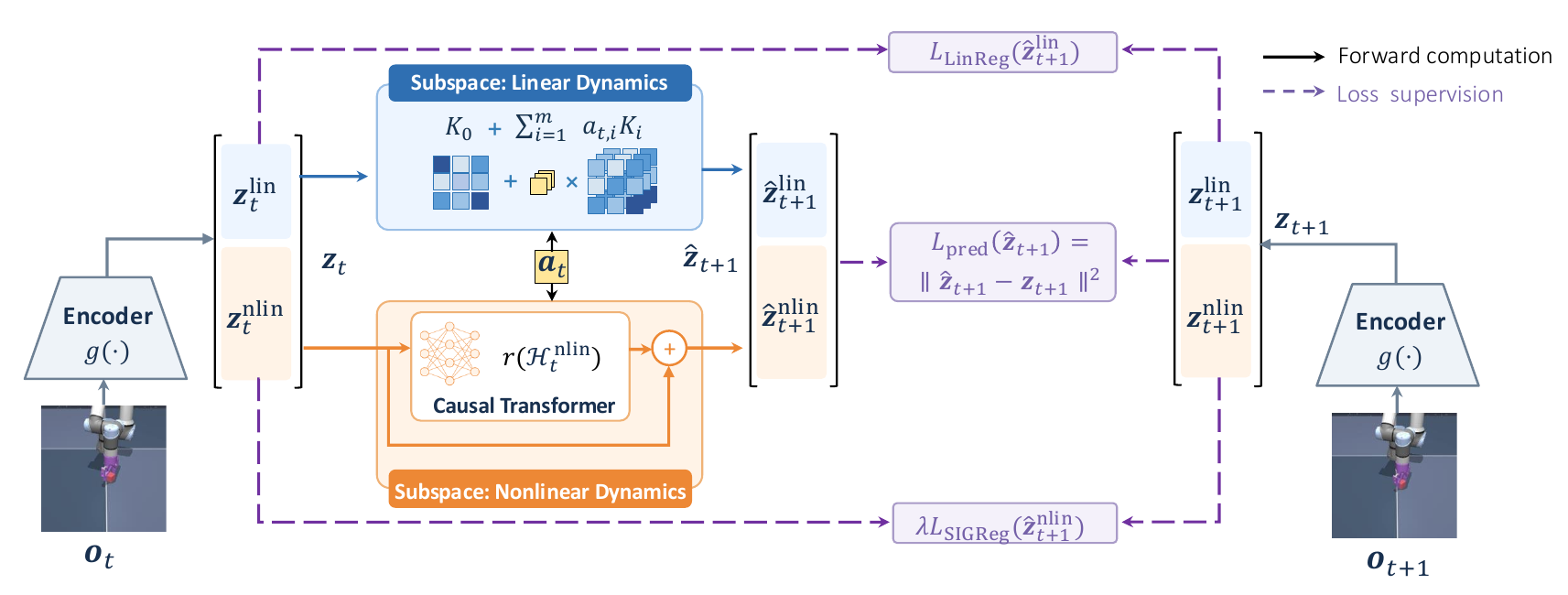}
    \caption{
    {\textbf{Framework of \ours.}
    The latent space is partitioned  into two  subspaces w.r.t. $\zref$ and $\zcont$ for the complementary linear and nonlinear dynamics, respectively. A linear map applies to $\zref$, whose evolution can be approximated by a linear transition  conditioned on action $\bm a$. A causal 
Transformer models  $\zcont$ with residual connections, using a history $\mathcal{H}^{\rm{nlin}}$ of recent latent states and actions for nonlinear temporal dependencies. The two subspaces are jointly optimized with a prediction loss and subspace-specific regularizers.
    }
    }
    \label{fig:framework}
\end{figure*}

In this work, we  propose \ours, a world model that jointly learns linear and nonlinear latent dynamics. A shared encoder produces a latent representation partitioned into two subspaces. An action-conditioned linear map advances the one subspace,  whose evolution can be approximated by a linear transition for each given action. A causal Transformer models the other, using a history of recent latent states and actions to capture nonlinear temporal dependencies. Together, the two branches form the latent representations through complementary predictive structures, {as illustrated in \cref{fig:framework}.} We jointly train both branches  with a prediction loss and subspace-specific regularization. 
 Variance and correlation regularization, namely LinReg, is applied to the linear subspace to maintain feature variation and reduce redundancy, while SIGReg is deployed to the nonlinear subspace to discourage representation collapse.

We evaluate \ours on four image-goal control environments: Push-T, TwoRoom, Reacher, and Cube. 
Across these environments, we achieve an average success rate {of 91.83\% for short-range goals and 69.50\% for distant goals.} In particular for  distant-goal planning, our average success rates exceed those of LeWM, SD-JEPA, and DINO-WM by 27.17\%, 24.67\%, and 10.50\%, respectively. Together with extensive further investigations and  ablations,  these results support our structured latent dynamics through complementary linear and nonlinear predictives for JEPAs.

Our contributions are summarized as follows:
\begin{itemize}
    \item We propose \ours, which structures the latent representation into  two jointly learned subspaces, assigned with an action-conditioned linear transition and a history-dependent nonlinear predictor, respectively.


    \item We develop a joint training objective that combines latent prediction with subspace-specific regularization to maintain representation diversity  and reduce redundancy among the linear coordinates.

    \item Experiments on four image-goal control
    environments  demonstrate our substantially higher average planning success rates for both short-range and distant goals, further supported by extensive     discussions and ablations.

\end{itemize}

\section{Related Work}

\paragraph{Latent World Models.} Latent world models support decision-making by predicting the effects of actions in a compact representation space~\cite{ha2018worldmodels,lecun2022path}. Their use in control includes online planning and policy learning from imagined experience. PlaNet~\cite{hafner2019planet} evaluates action sequences through latent rollouts, while MuZero~\cite{Schrittwieser_2020muzero} uses a learned model with reward, policy, and value predictions for tree search. The world model  proposed by Ha and Schmidhuber~\cite{ha2018worldmodels} provides representations for a compact controller, and the Dreamer family trains actor--critic policies using imagined trajectories~\cite{hafner2020dreamer,hafner2021dreamerv2,hafner2025dreamerv3}. Combining planning with learned value estimates, TD-MPC~\cite{hansen2022tdmpc} uses short-horizon latent rollouts and a terminal value function, and TD-MPC2~\cite{hansen2024tdmpc2scalablerobustworld}  extends this approach to multitask control. In JEPAs, it considers image-goal planning with a world model learned from reward-free offline trajectories. At test time, candidate actions are evaluated by comparing predicted latent states with the latent representation of the goal image.

\paragraph{Joint-Embedding Predictive Models.} JEPA architectures learn representations by predicting  the target latents~\cite{assran2023ijepa,bardes2024vjepa}. For action-conditioned world modeling, an important design choice is how the representations used for prediction are obtained. DINO-WM~\cite{zhou2025dinowm} trains a  predictor over spatial patch embeddings from a  pretrained strong visual encoder DINOv2.  PLDM~\cite{sobal2025pldm} instead jointly trains the encoder and  predictor. LeWM~\cite{maes2026lewm} considers  action-conditioning   and pioneers a stable end-to-end  framework  using   SIGReg \cite{balestriero2025lejepa}  to prevent representation collapse,
serving  strong performance.  
MotionJEPA \cite{karmann2026motionjepa} introduces an auxiliary frame-difference prediction objective to retain changing visual features, while SD-JEPA \cite{thil2026sdjepa} organizes representations for the progression and the content.  JEPA-Anything~\cite{cui2026jepaanything} factorizes the latents into decomposed  worlds across  domains of vision, biology, clinics, etc. These works primarily highlight the potentials of latent representation optimization, whereas the dynamics explicitly driving  future predictions are not yet explored, which we aim to address  by organizing latent coordinates according to their predictive mechanisms with complementary linear and nonlinear dynamics.

\paragraph{Latent Dynamics.}
Imposing structure on latent transitions provides an inductive bias for learning dynamics that can be used for control. One approach learns representations that support local linear models. Embed to Control~\cite{watter2015embed} learns locally linear latent dynamics for control from images, while SOLAR~\cite{zhang2019solar} fits local linear dynamics and quadratic cost models for iterative policy improvement. Other approaches can deploy Koopman representations, in which nonlinear dynamics  of observable are mapped to be descriptive through  linear evolutions \cite{williams2015data,brunton2016koopman,lusch2018deep}. Controlled formulations extend this methodology into action-conditioned operators \cite{haseli2026modeling}. Koopman Dreamer~\cite{li2026koopman} incorporates spectral constraints into the autonomous component of its latent dynamics to support imaginary rollouts for policy learning. Our \ours applies an action-conditioned linear transition to a learned subset of latent coordinates, and jointly models the remaining coordinates with a nonlinear predictor. Both subspaces contribute to the latent rollouts and goal-distance objective used for planning.

\section{Preliminaries}
\label{sec: pre}

We consider {image-goal} planning from {pixel-level} observations in JEPAs. Let \(\mathcal{O}\) and \(\mathcal{A}\) denote the spaces of observations and actions, respectively. At time step \(t\), {we have the observation \(\bm o_t \in \mathcal{O}\) in the world (environment), where  an action \(\bm a_t \in \mathcal{A}\) is taken and leads to the future observation \(\bm o_{t+1}\) in the next time step}. Given an initial observation \(\bm o_0\) and a goal image \(\bm o_g\), the task is to reach the goal through repeated planning over action sequences of length $H_p$. {A JEPA world model is learned} offline from a  set \(\mathcal{D}\) of trajectories {including observations and actions, i.e.,} \((\bm o_{0:T}, \bm a_{0:T-1})\), without rewards or task annotations.

{With the learned encoder and predictor, or namely the world model, the planner in JEPAs then evaluates a candidate action sequence
uses the world model to predict where the sequence ends and compares the predictions w.r.t. the goal in the latent space. We in this work focus on the modeling and optimization of such dynamics in this latent space.} 
Specifically, an encoder $g(\cdot)$ maps each observation
$\bm o_t$ to a latent state $\boldsymbol{z}_t=g(\bm o_t)$.
The history $\mathcal{H}_t= (\boldsymbol{z}_i,\boldsymbol{a}_i)_{i=\max(0,t-H+1)}^t$ contains up to $H$ latent--action pairs. The predictor estimates
the next latent state as $\widehat{\boldsymbol{z}}_{t+1}=f(\mathcal{H}_t)$.
Using $\boldsymbol{z}_{t+1}=g(\bm o_{t+1})$ as the target,
 the one-step prediction loss is given by
\begin{equation}
    L_{\mathrm{pred}}
    =
    \left\lVert
    \widehat{\boldsymbol{z}}_{t+1}-\boldsymbol{z}_{t+1}
    \right\rVert_2^2.
\label{eq:pred}
\end{equation}

{To prevent the representation collapse, the prediction loss in JEPAs is also combined with the objective that further optimize the latents, e.g.,  SIGReg \cite{balestriero2025lejepa}  in LeWM \cite{maes2026lewm}:}
\begin{equation}\label{eq:lewm}
 {L}
    =
    L_{\rm {pred}} + \lambda\,{L}_{\mathrm{SIGReg}}(\boldsymbol{z}),
\end{equation}
where $\lambda>0$ is the regularization constant and $\boldsymbol{z}$ denotes the collection
of latent samples used to compute the statistics within the regularization terms.
More specifically, the SIGReg objective is formulated as 
$L_{\mathrm{SIGReg}}(\boldsymbol{z})
=\frac{1}{M}\sum_{m=1}^{M}
\mathcal{T}(\{\boldsymbol{z}_j^\top\boldsymbol{u}^{(m)}\}_j)$,
where $j$ indexes the samples,
$\boldsymbol{u}^{(m)}\in\mathbb{S}^{d-1}$ is a randomly sampled
unit direction, and $\mathcal{T}(\cdot)$ is the univariate Epps--Pulley
statistic~\cite{epps1983test}. This objective encourages the projected samples
to follow isotropic Gaussian distributions for representation informativeness.

\section{Method}
\label{sec:method}
We introduce \ours, a latent world model that organizes latent representations into two subspaces capturing complementary linear and nonlinear dynamics, with details presented as follows. \cref{fig:framework} illustrates the architecture.  \cref{subsec: model} describes the two predictive branches and their connections across time steps. \cref{subsec: train} presents the joint training objective and the regularization specific to each subspace. \cref{subsec: plan} explains how the learned model supports planning through latent rollouts and model predictive control. Algorithm~\ref{alg:training}  summarizes the  training  procedures.

\subsection{Modeling of Linear and Nonlinear Dynamics}
\label{subsec: model}

For image-goal planning, the world model rolls out future latent states under candidate actions. LiND-JEPA partitions the latent representation into two subspaces with distinct predictive structures. An action-conditioned linear transition updates one subspace, while a history-conditioned nonlinear predictor produces an additive update for the other. The two branches are trained jointly with subspace-specific regularization, and their outputs are concatenated to form the predicted next state.

Let $\boldsymbol{z}_t = g(\bm o_t) \in \mathbb{R}^d$ the latent representation  from  the encoder. We structure the coordinates into two partitions:
$\boldsymbol{z}_t =
[\boldsymbol{z}_t^{\mathrm{lin}}; \boldsymbol{z}_t^{\mathrm{nlin}}]$,
with $d_{\mathrm{lin}} + d_{\mathrm{nlin}} = d$.
For the first subspace branch, we consider a transition dynamic that is linear w.r.t. the current
latent state $\boldsymbol{z}_t^{\mathrm{lin}}$ conditioned on action $\bm a_t$:
\begin{equation}
\begin{aligned}
    \widehat{\boldsymbol{z}}_{t+1}^{\mathrm{lin}}
    &= K(\boldsymbol{a}_t)\boldsymbol{z}_t^{\mathrm{lin}}, \\
    K(\boldsymbol{a}_t)
    &= K_0 + \sum\nolimits_{i=1}^{m} a_{t,i} K_i ,
\label{eq:linear}
\end{aligned}
\end{equation}
where $m$ is the action dimension, $a_{t,i}$ is the $i$-th action
component, and $K_0,\ldots,K_m \in
\mathbb{R}^{d_{\mathrm{lin}}\times d_{\mathrm{lin}}}$ are learnable
matrices. We initialize $K_0$ to the identity matrix and each $K_i$  to zero for $i=1,\ldots,m$. Thus, this branch simply copies its input at
initialization. 
During training, the prediction loss encourages the encoder to learn the corresponding latents,  for which an action-conditioned linear predictor $K(\bm a_t)\bm z_t^{\mathrm{lin}}$ can approximate  the next-step representation $\boldsymbol{z}_{t+1}^{\mathrm{lin}}$.

For the nonlinear subspace, a causal Transformer $r(\cdot)$
predicts an increment from recent latent states and actions.
Let
$\mathcal{H}_t^{\mathrm{nlin}}
=(\boldsymbol{z}_i^{\mathrm{nlin}},
   \boldsymbol{a}_i)_{i=\max(0,t-H+1)}^t$
denote the history of up to $H$ pairs.
We add the predicted increment to the current latent state:
\begin{equation}
    \widehat{\boldsymbol{z}}_{t+1}^{\mathrm{nlin}}
    =
    \boldsymbol{z}_t^{\mathrm{nlin}}
    +r(\mathcal{H}_t^{\mathrm{nlin}}).
\label{eq:nonlinear}
\end{equation}
The additive update carries the current state forward while
modeling its change from recent history.
We concatenate the two branch outputs to obtain
$\widehat{\boldsymbol{z}}_{t+1}
=[\widehat{\boldsymbol{z}}_{t+1}^{\mathrm{lin}};
  \widehat{\boldsymbol{z}}_{t+1}^{\mathrm{nlin}}]$.

\subsection{Training}
\label{subsec: train}
We jointly train the encoder, the action-conditioned linear predictor, and the nonlinear predictor. The prediction loss $L_{\mathrm{pred}}$ supervises the concatenated prediction with the target latent representation at the next time step, encouraging both branches to capture the temporal evolution within their respective subspaces. To maintain informative representations in both subspaces, we complement this loss with subspace-specific regularization.

For the linear coordinates, we aim to maintain variation
and reduce redundancy while allowing their scale to change
under the learned transition. Motivated by the variance and covariance terms of VICReg~\cite{bardes2022vicreg}, we combine a
variance penalty with a correlation penalty.
Let $\sigma^2(\boldsymbol{z}^{\mathrm{lin}})$ denote the vector of  coordinate-wise variances pooled over the history and batch dimensions. The variance regularizer is cast as
\begin{equation}
    L_{\mathrm{var}}(\boldsymbol{z}^{\mathrm{lin}})
    =
    \frac{1}{d_{\mathrm{lin}}}
    \bigg \lVert
        \left[
            \gamma -
            \sqrt{\sigma^2(\boldsymbol{z}^{\mathrm{lin}}) + \epsilon}
        \right]_+
   \bigg\rVert_2^2,
\end{equation}
where $[\cdot]_+$ applies $\max(\cdot,0)$ element-wisely, $\gamma$ is the  threshold of standard deviation and set to 1,
and $\epsilon = 10^{-4}$ is for numerical stability.
This term discourages each latent coordinate from collapsing to
a constant across samples.

To reduce redundancy among the linear coordinates, let $X_h\in\mathbb{R}^{B\times d_{\mathrm{lin}}}$ collect the linear latent vectors at history position $h$ as rows. We obtain $\widetilde{X}_h$ by centering each column and dividing it by its batch standard deviation. With $C_h=B^{-1}\widetilde{X}_h^\top\widetilde{X}_h$, the correlation  regularizer is given by
\begin{equation}
    L_{\mathrm{corr}}(\boldsymbol{z}^{\mathrm{lin}})
    =
    \frac{
    \sum_{h=1}^{H}
    \sum_{1\leq i\neq j\leq d_{\mathrm{lin}}}
    (C_h)_{ij}^2
    }{H d_{\mathrm{lin}}(d_{\mathrm{lin}}-1)},
\end{equation}
which penalizes off-diagonal correlations  between distinct latent coordinates at each history position.
We combine the two terms into our namely LinReg regularization:
\begin{equation}\label{eq:linreg}
\begin{aligned}
    L_{\mathrm{LinReg}}(\boldsymbol{z}^{\mathrm{lin}})
    ={}
    \lambda_{\mathrm{var}}
    L_{\mathrm{var}}(\boldsymbol{z}^{\mathrm{lin}}) +
    \lambda_{\mathrm{corr}}
    L_{\mathrm{corr}}(\boldsymbol{z}^{\mathrm{lin}}).
\end{aligned}
\end{equation}

For the nonlinear subspace, we use
$L_{\mathrm{SIGReg}}(\boldsymbol{z}^{\mathrm{nlin}})$
to encourage the projected latent distributions to match
a standard Gaussian, similar to LeWM. This discourages representation collapse
and maintains variation across projection directions, while
the prediction loss trains the nonlinear dynamics.
Thus, our overall objective is formulated as
\begin{equation}\label{eq:full:loss}
\begin{aligned}
    L_{\mathrm{all}}
    ={}
    L_{\mathrm{pred}} 
    + L_{\mathrm{LinReg}}(\boldsymbol{z}^{\mathrm{lin}}) +
    \lambda
    L_{\mathrm{SIGReg}}(\boldsymbol{z}^{\mathrm{nlin}}),
\end{aligned}
\end{equation}
where $\lambda_{\mathrm{var}}$, $\lambda_{\mathrm{corr}}$,
and $\lambda$ are  regularization coefficients.

\begin{algorithm}[t]
\caption{Training with \ours}
\label{alg:training}
\small
\begin{algorithmic}[1]
\item[\textbf{Given:}] {Offline trajectories} $\mathcal{D}$,
 batch size $B$, 
history length $H$, 
 regularization coefficients $\lambda_{\mathrm{var}},
         \lambda_{\mathrm{corr}},\lambda$.
\item[\textbf{Learn:}]  Parameters  $\bm \theta=\{\texttt{params}\{g(\cdot),r(\cdot)\},K_0,\ldots,K_m\}$.
\State Initialize  
       $\bm \theta$
       with $K_0=\mathbf I_{d_{\mathrm{lin}}}$ and $K_i=\mathbf 0$ for $i=1,\ldots,m$.
\While{training budget remains}
    \State Sample $B$ trajectory segments  with $H$ input
           positions, actions, and corresponding next-observation targets from $\mathcal{D}$;
    \State Encode observations with $g(\cdot)$ and
           construct history $\mathcal{H}_t$;
    \State Predict two subspaces with \cref{eq:linear}--\cref{eq:nonlinear}: 
$\widehat{\boldsymbol{z}}_{t+1}
= [\widehat{\boldsymbol{z}}_{t+1}^{\mathrm{lin}};
   \widehat{\boldsymbol{z}}_{t+1}^{\mathrm{nlin}}]$;
    \State Compute prediction loss $L_{\mathrm{pred}}$, and  regularizers $L_{\mathrm{LinReg}}$
           using \cref{eq:linreg}
     and
           $L_{\mathrm{SIGReg}}$ referring to \cref{eq:lewm};
    \State Update $\bm \theta$ by minimizing the full loss $L_{\mathrm{all}}$
           in \cref{eq:full:loss}.
\EndWhile
\State \Return  \ours model with the learned parameters $\bm \theta$.
\end{algorithmic}
\end{algorithm}

\subsection{Planning}
\label{subsec: plan}
We use the learned encoder and predictive model for image-goal planning within model predictive control (MPC) \cite{mayne2000mpc} following prior work under JEPA frameworks \cite{maes2026lewm,zhou2025dinowm}. At each planning step $t$, the observation $\bm o_t$ and the goal image $o_g$ are encoded as $\boldsymbol{z}_t = g(\bm o_t)$ and $\boldsymbol{z}_g = g(\bm o_g)$, respectively. Both representations contain the linear and nonlinear subspaces defined above. {Recent observations are encoded in the same way and paired with the actions to form the history  $\mathcal{H}_t^{\mathrm{nlin}}$ used by the nonlinear predictor.}

For each candidate action sequence $\boldsymbol{a}_{t:t+H_p-1}$ with   the planning
horizon of $H_p$, we initialize the rollout with $\widehat{\boldsymbol{z}}_t = \boldsymbol{z}_t$ and recursively apply both predictive branches in rollouts. At each rollout step $k=0,\ldots,H_p-1$, the linear branch computes $\widehat{\boldsymbol{z}}_{t+k+1}^{\mathrm{lin}} = K(\boldsymbol{a}_{t+k}) \widehat{\boldsymbol{z}}_{t+k}^{\mathrm{lin}}$, while the nonlinear branch computes $\widehat{\boldsymbol{z}}_{t+k+1}^{\mathrm{nlin}} = \widehat{\boldsymbol{z}}_{t+k}^{\mathrm{nlin}} + r(\widehat{\mathcal{H}}^{\mathrm{nlin}}_{t+k})$, where $\widehat{\mathcal{H}}_{t+k}^{\mathrm{nlin}}$ contains the available  history as in \cref{eq:nonlinear}, retaining at most $H$ pairs and ending up with $(\widehat{\boldsymbol{z}}^{\mathrm{nlin}}_{t+k},\boldsymbol{a}_{t+k})$. We concatenate two branch outputs to complete the latent prediction $\widehat{\boldsymbol{z}}_{t+k+1}$ and use it for the next rollout step.

The planner seeks an action sequence whose predicted terminal representation is close to the goal representation:
\begin{equation}\label{eq:planning}
    \boldsymbol{a}^{*}_{t:t+H_p-1}
    =
    \underset{
        \boldsymbol{a}_{t:t+H_p-1} \in \mathcal{A}^{H_p}
    }{\arg\min}
    \left\lVert
        \widehat{\boldsymbol{z}}_{t+H_p}
        - \boldsymbol{z}_g
    \right\rVert_2^2,
\end{equation}
where $H_p$ is the planning horizon and $\widehat{\boldsymbol{z}}_{t+H_p}$ depends on the candidate action sequence through the recursive rollout steps. The distance is computed over the predicted full latents, constituted by the representations with our two subspaces.
For this   optimization problem, we take the commonly used solver of cross-entropy method (CEM)~\cite{rubinstein1999crossentropy}.

\section{Theoretical Analysis}
\label{sec:theoretical-analysis}
We analyze the one-step prediction generalization ability of our structured latent dynamics, given  finite $N$ samples . 
We compare our \textit{s}tructured model \ours using subscript ``$s$'' with a dimension-matched
\textit{u}nstructured comparator using subscript ``$u$''. Let $\mathcal J_s$ and $\mathcal J_u$
denote their joint function classes of encoder $g$ and predictor $f$, respectively.
Both use the same encoder class, latent split, and
predictor class for the nonlinear space $\bm z^{\mathrm{nlin}}$ in \cref{eq:nonlinear}.
For the linear subspace $\zref$, the comparator remains using a nonlinear predictor, while ours adopts a linear one. We assume that it can
realize every admissible update in \cref{eq:linear} with the encoder
and the predictor for $\bm z_t^{\mathrm{nlin}}$ fixed, which gives
$\mathcal J_s\subseteq\mathcal J_u$.


Let $S=(X_1,\ldots,X_N)$ denote a set of $N$ data sequences, where  each $X_i$ represents an individual training sample.
We assume that these sequences are independently and identically distributed according to $P_X$, i.e., $S\sim P_X^N$.
The classes $\mathcal J_s$ and $\mathcal J_u$ are fixed before sampling.
Assume the one-step  latent
prediction loss satisfies $0\leq\ell(e;X)\leq M_\ell<\infty$ for every
$e=(g,f)$ in either class and every $X$ in the support of $P_X$.
For $q\in\{s,u\}$ and $e\in\mathcal J_q$, write $\mathcal R_q(e)=\mathbb E_{X\sim P_X}[\ell(e;X)]$
for the population risk and $\widehat{\mathcal R}_S(e)=N^{-1}\sum_{i=1}^N\ell(e;X_i)$
for the empirical risk. Let $\mathcal R_q^\star=\inf_{e\in\mathcal J_q}\mathcal R_q(e)$
and let $\widehat{\mathfrak R}_q$ denote the empirical Rademacher complexity
of the loss class induced by $\mathcal J_q$~\cite{bartlett2002rademacher}.

\begin{theorem}[One-step generalization]
Let $e^\star_q$ minimize the empirical prediction risk in
$\mathcal J_q$ on the given $N$ i.i.d. samples.
For class $q\in\{s, u\}$, given any $\delta\in(0,1)$, with probability at least $1-\delta$,
both classes satisfy
\begin{equation}
\label{eq:theory-risk-bound}
\mathcal R_q(e^\star_q)\leq
\mathcal R_q^\star+4\widehat{\mathfrak R}_q
+6M_\ell\sqrt{\frac{\log(8/\delta)}{2N}}.
\end{equation}
\end{theorem}

We consider the prediction loss split as $\ell=\ell^{\mathrm{lin}}+\ell^{\mathrm{nlin}}$,
with one term for each latent subspace. Over the structured class $q:=s$, i.e.,
$\mathcal J_s$, let $\widehat{\mathfrak R}_{\mathrm{nlin}}$ be the empirical
Rademacher complexity of $\ell^{\mathrm{nlin}}$ w.r.t. the nonlinear subspace, and let
$B_{d_{\mathrm{lin}}}(\varepsilon)$ bound that of $\ell^{\mathrm{lin}}$ w.r.t. the linear subspace. We have  following lemma that bounds the empirical Rademacher complexity for the structured class $\mathcal{J}_s$. 
\begin{lemma}\label{lemma}
For any covering radius $\varepsilon>0$, the  complexity 
$\widehat{\mathfrak R}_s$ of the structured \ours satisfies
\begin{equation}
\label{eq:theory-split-complexity}
\begin{aligned}
\widehat{\mathfrak R}_s
&\leq \widehat{\mathfrak R}_{\mathrm{nlin}}+B_{d_{\mathrm{lin}}}(\varepsilon),\\
B_{d_{\mathrm{lin}}}(\varepsilon)
&=\frac{M_\ell}{2}\bigg(\varepsilon+
\frac{\sqrt{\rho_{\mathrm{lin}}}+\sqrt{2\log C_{\mathrm{lin}}(\varepsilon)}}
     {\sqrt N}\bigg),
\end{aligned}
\end{equation}
where $\rho_{\mathrm{lin}}$ is the maximum dimension of the spaces spanned by the all-ones vector $\bm{1}$ and the loss vectors on $S$ from all admissible linear predictors for each fixed encoder.
$C_{\mathrm{lin}}(\varepsilon)$ is the covering number of  orthogonal projectors onto these spans for approximation within $\varepsilon$ in
operator norm \cite{bartlett2002rademacher}.
\end{lemma}

\begin{corollary}
For the $k$ selected samples, each with two prescribed  prediction losses on $\zref$, let $\beta>0$ be a common lower bound on their respective gaps.
Let $\widehat{\mathfrak R}_{\mathrm{nlin}}(g)$ denote the empirical Rademacher complexity of the loss  w.r.t. the structured nonlinear subspace with a prespecified  encoder $g$ and the varying predictor $f$. 
Under the assumptions of Appendix~\ref{app:corollary}, if
\begin{equation}
\frac{k\beta}{2N}
>
\inf_{\varepsilon>0} B_{d_{\mathrm{lin}}}(\varepsilon)
+\frac{\mathcal R_s^\star-\mathcal R_u^\star}{4}
+\bigl(\widehat{\mathfrak R}_{\mathrm{nlin}}
       -\widehat{\mathfrak R}_{\mathrm{nlin}}(g)\bigr),
\label{eq:corollary}
\end{equation} 
then for some $\varepsilon>0$, it yields an explicit upper bound on one-step
population risk that is strictly smaller than the corresponding
bound for the unstructured comparator.
\end{corollary}

When \cref{eq:corollary} holds, \ours has a strictly tighter upper bound on one-step population risk than the unstructured model. We provide detailed proofs and discussions in Appendix~\ref{app:theory-proofs}.


\section{Experiments}

We evaluate LiND-JEPA through three questions.
\textbf{(1)} How well does it perform image-goal planning across
environments and goal distances?
\textbf{(2)} How do the predictive structure and subspace-specific
regularization contribute to its performance?
\textbf{(3)} What properties of the learned representations and
executions characterize its planning behavior?
We first describe the experimental setup, then address these questions through benchmark comparisons, controlled ablations, and representation and trajectory analyses. {Detailed setups and additional results are provided in Appendix~\ref{app:implementation-details}.}

\paragraph{Experimental Setup.}
We evaluate on four image-goal control environments:
PushT~\cite{hi2023diffusionpolicy}, TwoRoom~\cite{sobal2025pldm}, DMControl Reacher~\cite{tassa2018dmc}, and OGBench-Cube~\cite{park2025ogbench}  used in LeWM~\cite{maes2026lewm} and SD-JEPA \cite{thil2026sdjepa}.
Each task provides an initial observation and a goal image, and the agent selects actions to reach the goal within an execution budget. Models are trained on offline trajectories without reward annotations.
We consider two evaluation protocols. The short-range protocol uses a start--goal offset of 25 trajectory steps and an execution budget of 50 environment steps.
The distant-goal protocol increases these values to 100 and 200 steps, respectively. Both protocols use an MPC horizon of 25 environment steps and the same trained checkpoints.

\begin{table*}[t]
\centering
\small
\setlength{\tabcolsep}{4pt}
\renewcommand{\arraystretch}{1.12}
\caption{Short-range image-goal planning success (\%).
The start--goal offset is 25 trajectory steps and the execution
budget is 50 environment steps. Results are mean $\pm$ standard
deviation across three training seeds, each evaluated on the
same 50 tasks per environment. Avg.\ is the unweighted mean
across environments. Bold indicates the highest mean.}
\label{tab:planning-offset25}
\begin{tabular*}{0.68\textwidth}{@{\extracolsep{\fill}}lccccc@{}}
\toprule
Method & PushT & TwoRoom & Reacher & Cube & Avg. \\
\midrule
LeWM
& $\mathbf{95.33\pm2.31}$
& $86.67\pm3.06$
& $80.00\pm2.00$
& $68.00\pm5.29$
& $82.50$ \\

SD-JEPA
& $93.33\pm3.06$
& $86.67\pm2.31$
& $78.67\pm5.03$
& $68.00\pm5.29$
& $81.67$ \\

DINO-WM
& $78.67\pm1.15$
& $\mathbf{100.00\pm0.00}$
& $71.33\pm5.03$
& $76.00\pm5.29$
& $81.50$ \\

\ours
& $93.33\pm1.15$
& $\mathbf{100.00\pm0.00}$
& $\mathbf{82.00\pm5.29}$
& $\mathbf{92.00\pm3.46}$
& $\mathbf{91.83}$ \\
\bottomrule
\end{tabular*}
\end{table*}

\begin{table*}[t]
\centering
\small
\setlength{\tabcolsep}{4pt}
\renewcommand{\arraystretch}{1.12}
\caption{Distant-goal planning success (\%).
The start--goal offset is 100 trajectory steps and the execution
budget is 200 environment steps. We use the same checkpoints as in
Tab.~\ref{tab:planning-offset25} and retain the same MPC horizon. Results are mean $\pm$ standard deviation
over three  seeds.}
\label{tab:planning-offset100}
\begin{tabular*}{0.68\textwidth}{@{\extracolsep{\fill}}lccccc@{}}
\toprule
Method & PushT & TwoRoom & Reacher & Cube & Avg. \\
\midrule
LeWM
& $\mathbf{16.00\pm4.00}$
& $22.00\pm5.29$
& $73.33\pm2.31$
& $58.00\pm2.00$
& $42.33$ \\

SD-JEPA
& $15.33\pm3.06$
& $22.00\pm2.00$
& $79.33\pm1.15$
& $62.67\pm5.03$
& $44.83$ \\

DINO-WM
& $10.67\pm2.31$
& $64.67\pm5.03$
& $96.00\pm0.00$
& $64.67\pm6.43$
& $59.00$ \\

\ours
& $11.33\pm3.06$
& $\mathbf{94.00\pm2.00}$
& $\mathbf{100.00\pm0.00}$
& $\mathbf{72.67\pm1.15}$
& $\mathbf{69.50}$ \\
\bottomrule
\end{tabular*}
\end{table*}

We present main comparisons with LeWM~\cite{maes2026lewm}, SD-JEPA~\cite{thil2026sdjepa}, and DINO-WM~\cite{zhou2025dinowm}. LeWM and SD-JEPA jointly learn representations and predictive dynamics, while DINO-WM uses a pretrained DINOv2~\cite{oquab2024dinov2} encoder for strong visual capabilities. The methods use the same offline trajectories within each environment. LiND-JEPA is trained for 10 epochs with a batch size of $B=128$ and a latent dimension of $d=192$. At test time, it uses MPC with CEM to optimize candidate action sequences. We report results over three training seeds, evaluated on the same 50 tasks per environment for each protocol. \cref{tab:planning-offset25} and \cref{tab:planning-offset100} report mean success rates and standard deviations across seeds; the average column is the  mean across the four environments.

\subsection{Image-Goal Planning Performance}
\label{sec:planning_results}

\paragraph{Short-Range Goals.}
\cref{tab:planning-offset25} reports the results of short-range planning success.
\ours achieves the highest average success rate of
91.83\%, substantially outperforming the 82.50\% for LeWM, 81.67\% for
SD-JEPA, and 81.50\% for DINO-WM.
The largest improvement occurs on the OGBench Cube environment of 3D manipulation, where LiND-JEPA
reaches 92.00\%, in contrast to 76.00\% for DINO-WM and
68.00\% for both LeWM and SD-JEPA.
We also reach 100.00\% on TwoRoom and 82.00\% on Reacher, giving the best performance for these two environments.
On Push-T, LiND-JEPA obtains 93.33\%, matching SD-JEPA
and slightly falling below LeWM's 95.33\%.
The results show improvements in average planning success,
with gains that vary across environments.

\paragraph{Distant Goals.}
\cref{tab:planning-offset100} evaluates more distant goals
with a larger execution budget, while keeping the MPC horizon
fixed. \ours still achieves the highest average success rate of 69.50\%,
exceeding LeWM, SD-JEPA, and DINO-WM by 27.17\%, 24.67\%,
and 10.50\%, respectively.
The largest gain is on TwoRoom, where \ours reaches
94.00\%, compared with 22.00\% for both LeWM and SD-JEPA
and 64.67\% for DINO-WM which deploy strong visual encoder.
LiND-JEPA also maintains 100.00\% on Reacher as in \cref{tab:planning-offset100}  and achieves the advantageous performance of 72.67\%
on Cube.

The distant-goal Push-T remains challenging for all these evaluated models. \ours obtains
11.33\%, compared with 16.00\% for LeWM, 15.33\% for
SD-JEPA, and 10.67\% for DINO-WM.
Thus, the average improvement is driven primarily by
TwoRoom, Reacher, and Cube.
Reacher success is higher under the distant-goal protocol,
which also provides more execution steps. Since goal distance
and execution budget both change between protocols, this
comparison alone does not isolate the effect of either factor.

\subsection{Contributions of Model Components}
\label{sec:ablations}

We examine our design of the linear transition and subspace-specific
regularization, and also the allocation of latent dimensions in two subspaces.
The transition and regularization ablations are conducted
on Cube under both goal protocols, while the dimension
study covers all four environments under the short-range
protocol. Each configuration is evaluated with an  average  over three
training seeds.

\begin{table}[t]
    \centering
    \small
    \setlength{\tabcolsep}{3pt}
    \caption{Linear predictor ablation.
    Success rates are averaged over three training seeds for each planning
    protocol.} 
    \label{tab:linear_transition_ablation}
    \resizebox{\columnwidth}{!}{%
    \begin{tabular}{@{}llcc@{}}
        \toprule
        Variant & Linear Predictor & 
         Short-range & Distant-range \\
        \midrule
        \texttt{Identity}
            & $\zrefhat_{t+1} = \zref_t$
            & 84.67 & 60.67 \\
        \texttt{Action-Independent}
            & $\zrefhat_{t+1} = K_0\zref_t$
            & 82.67 & 61.33 \\
        \textbf{\ours}
            & $\zrefhat_{t+1} = (K_0 + \sum_i^m a_{t,i}K_i)\zref_t$
            & 92.00 & 72.67\\
        \midrule
        LeWM  & \textit{N/A} & 68.00 & 58.00 \\
        \bottomrule
    \end{tabular}%
    }
\end{table}

\begin{table}[t]
\centering
\small
\setlength{\tabcolsep}{3pt}
\caption{Nonlinear prediction ablation on Cube.
For each protocol, we average success rates (\%) over three training seeds.}
\label{tab:incremental-prediction-ablation}
 \resizebox{\columnwidth}{!}{%
\begin{tabular}{@{}llcc@{}}
\toprule
Variant & Nonlinear predictor & Short-Range  & Distant-Range  \\
\midrule
\texttt{w/o Residual} &  $\zconthat_{t+1}=r(\mathcal H_t^{\mathrm{nlin}})$
         & 80.00 & 68.00 \\
\textbf{\ours}  & $\zconthat_{t+1}=\zcont_t+r(\mathcal H_t^{\mathrm{nlin}})$
        & 92.00 & 72.67  \\
          \midrule
        LeWM  & $\widehat{\bm z}_{t+1}=f(\mathcal H_t)$ & 68.00 & 58.00 \\
\bottomrule
\end{tabular}
}
\end{table}

\paragraph{Action-Conditioned Linear Transition.}
\cref{tab:linear_transition_ablation} compares the linear transition of our \ours
in \cref{eq:linear} with two alternatives.
The \texttt{Identity} variant directly copies the linear latent
state, while the \texttt{Action-Independent} variant learns a 
matrix $K_0$ for linearity without action conditioning.
Our action-conditioned transition achieves
92.00\% and 72.67\% success rates under the short-range and
distant-goal protocols, respectively.
The corresponding results are 84.67\% and 60.67\% for
\texttt{Identity}, and 82.67\% and 61.33\% for \texttt{Action-Independent}.
These demonstrate that our leveraged action-conditioned variant performs the best under both
protocols, supporting the significance of actions to model the temporal evolution along  the coordinates in the linear subspace.

\paragraph{Nonlinear Update Formulation.}
In \cref{tab:incremental-prediction-ablation}, we assess the effect of incremental prediction upon residual connections of nonlinear latents. We ablate to compare our additive update with directly predicting the next nonlinear latent states, named as the \texttt{w/o Residual} variant. The additive formulation improves success from 80.00\% to 92.00\% for short-range goals and from 68.00\% to 72.67\% for distant goals on Cube. These results support using the current nonlinear state as a reference when predicting its temporal evolution. Note that our  variant   \texttt{w/o Residual} still outperforms LeWM  under both protocols, further supporting   the effectiveness of  our structured latent modeling.

\paragraph{Subspace-Specific Regularization.}
\cref{tab:regularization-ablation}~ compares our subspace-specific regularization with \texttt{Full SIGReg}, which applies SIGReg to the complete latent representation of our two-branch model. Our \ours instead applies LinReg to the linear coordinates and SIGReg to the nonlinear coordinates. This improves the short-range success rate from 82.00\% to 92.00\% and distant-goal success from 66.67\% to 72.67\%, corresponding to gains of 10.00\% and 6.00\%. These results provide supportive evidence using different regularizers for the structured two subspaces.

\begin{table}[t]
\centering
\small
\setlength{\tabcolsep}{3pt}
\renewcommand{\arraystretch}{1.12}
\caption{Subspace-specific regularization ablation.
Success rates (\%) are averaged over three training seeds on  Cube.}

\label{tab:regularization-ablation}
\resizebox{\columnwidth}{!}{%
\begin{tabular}{@{}llcc@{}}
\toprule
Variant & Regularizer & Short-range & Distant-range \\
\midrule
\texttt{Full SIGReg} & $L_\mathrm{SIGReg}([\zref;\zcont])$
      & 82.00 & 66.67 \\
\textbf{\ours} & $L_\mathrm{LinReg}(\zref) + L_\mathrm{SIGReg}(\zcont)$
      & 92.00 & 72.67 \\
\bottomrule
\end{tabular}%
}
\end{table}

\begin{figure}[t]
    \centering
    \includegraphics[width=0.85\columnwidth]{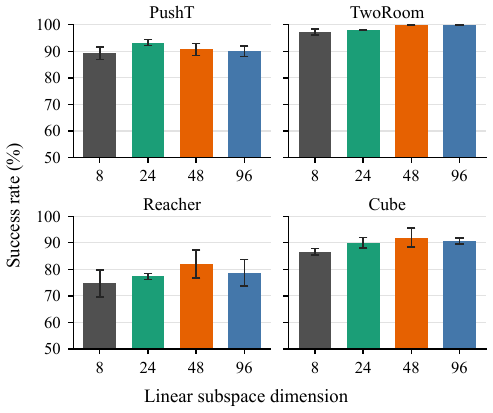}
    \caption{Linear subspace dimension ablation at a 25-step goal offset.
    Bars show mean planning success over three training seeds, with error
    bars indicating $\pm$ one sample standard deviation.}
    \label{fig:linear-dim-offset25}
\end{figure}

\paragraph{Latent Dimension Split.}
We vary the dimension of the linear subspace, i.e., $d_{\mathrm{lin}}\in\{8,24,48,96\}$, with the given total latent dimension of $d=192$, such that the dimension of the nonlinear subspace is accordingly adjusted as  $d_{\mathrm{nlin}}=d-d_{\mathrm{lin}}$. \cref{fig:linear-dim-offset25}~shows that the preferred allocation varies across environments. Push-T achieves its highest success at $d_{\mathrm{lin}}=24$, while Reacher and Cube peak at $d_{\mathrm{lin}}=48$. TwoRoom reaches 100\% at both 48 and 96 linear dimensions, suggesting that an easier task (e.g., TwoRoom)  posing relatively simple dynamics can achieve good performance with a larger linear subspace and a correspondingly smaller nonlinear subspace.  These results indicate that allocating capacity between the two predictive structures affects environments differently. Among the evaluated setups, our performance in general remains high and varies modestly.

\begin{figure}[t]
    \centering
    \includegraphics[width=0.85\linewidth]{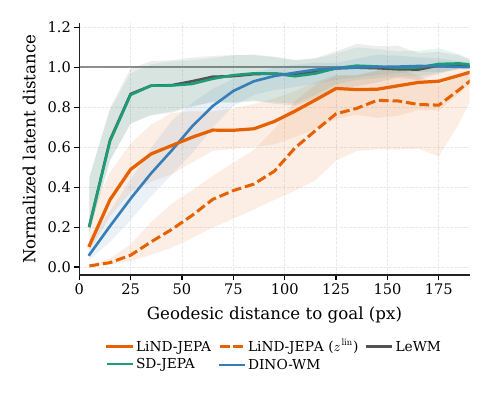}
    \caption{Relation between latent and physical goal distances on TwoRoom. We plot normalized squared latent goal distance against physical geodesic distance using encoded observations from 50 expert trajectories. The linear-subspace curve \ours($\zref$) uses only the coordinates $\boldsymbol{z}^{\mathrm{lin}}$ in  distance computation.}
    \label{fig:distance-saturation}
\end{figure}

\begin{figure*}[ht]
    \centering
    \includegraphics[width=0.7\textwidth]
    {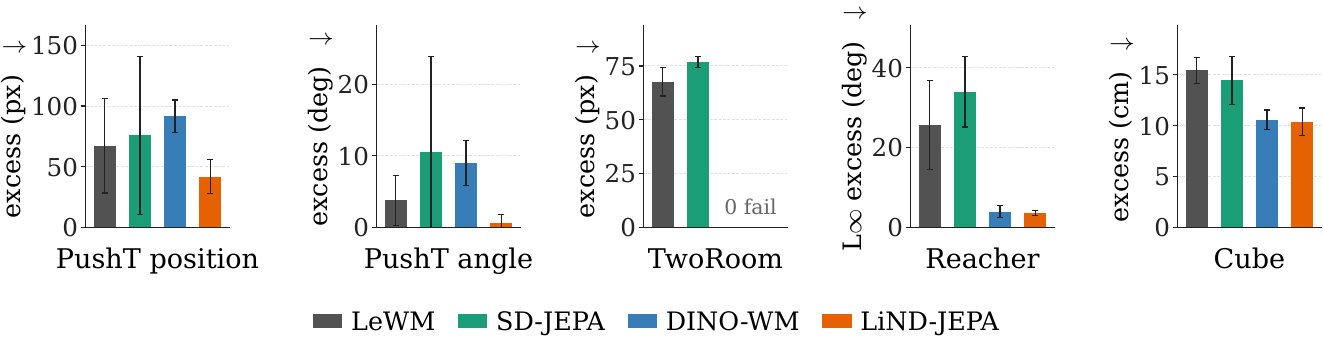}
    \caption
    {
    Terminal goal gaps on failed  trials under the short-range goal protocol. Smaller values indicate closer proximity to the corresponding success threshold. Each method is evaluated on its own failed trials. Methods with no failed trials have no defined  statistic.
    }
    \label{fig:offset25-failure-excess}
\end{figure*}

\begin{figure*}[ht]
    \centering
    \includegraphics[width=0.95\textwidth]
    {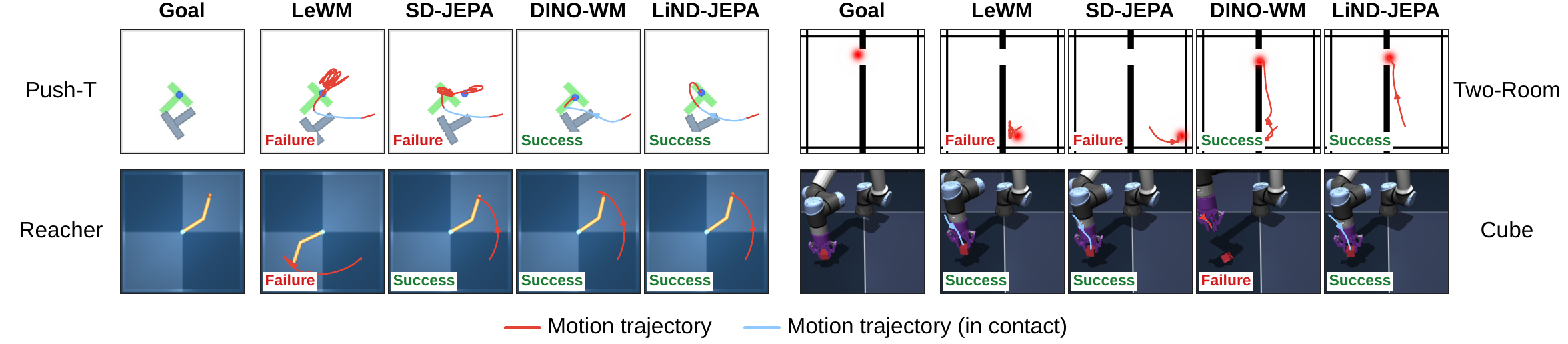}
    \caption
    {
    Exemplified visualization on the planning trajectories for the compared baselines and \ours across four environments. Each example shows the  goal, executed trajectory, and final success or failure outcome of the executed task.
    }
    \label{fig:visualization}
\end{figure*}

\subsection{Representation and Planning Behavior}
\label{sec:representation_analysis}

We here further explore the representation and planning behavior of our \ours in terms of the goal-distance measure  in the learned latent space used by the
planner and the terminal errors of unsuccessful trials, together with exemplification on executed trajectories.

\paragraph{Latent and Physical Goal Distances.}
The planning objective in ~\cref{eq:planning} uses the latent distance to
score candidate action sequences.
\cref{fig:distance-saturation} examines how this measure
relates to the physical goal proximity in pixels on TwoRoom.
Using frozen encoders from three training seeds, we encode
every fifth frame from 50 expert trajectories under the
distant-goal setting.
We plot the normalized squared latent distance
$\|\boldsymbol{z}_t-\boldsymbol{z}_g\|_2^2/
 \|\boldsymbol{z}_0-\boldsymbol{z}_g\|_2^2$
against physical geodesic distance to the goal.

The curves for LeWM, SD-JEPA, and DINO-WM become flatter
as physical distance increases, and get saturated. In these regions with saturation, states
at different physical distances receive similar latent
goal distances, making it difficult to distinguish the true distance to the goal.
\ours maintains a clearer association over a wider
range, particularly in its linear coordinates $\zref$.
This observation suggests that our learned representation
provides a more informative measure of goal proximity, as evaluated on TwoRoom trajectories.
This analysis uses encoded observations and characterizes
representation geometry; the prediction accuracy during model
rollouts remains a separate factor in planning performance.

\paragraph{Final Gap on Failed Trials.}
We  examine the trials that do not reach the success criterion within the short-range execution
budget, as shown in \cref{fig:offset25-failure-excess}. For each reported task metric, we measure the terminal
distance relative to its success threshold. Smaller gaps indicate that unfinished trials terminate closer to satisfying the corresponding success criterion. Each method's average is computed over its own failed trials, so the compared subsets can differ. On TwoRoom, LiND-JEPA and DINO-WM have no failed trials, and their failure-gap statistics are therefore undefined. 

\ours shows smaller terminal gaps on the reported metrics for all tasks, with a pronounced difference on Push-T. This provides additional context for its Push-T performance, where its success rate is slightly below that of LeWM. These results in \cref{fig:offset25-failure-excess} describe the outcomes of unsuccessful executions alongside the success rates.

\paragraph{Trajectory Visualization.}
\cref{fig:visualization} exemplies the executions for LeWM, DINO-WM, SD-JEPA and our \ours across the four environments. These visualizations show the paths taken and their final
success or failure outcomes. They provide qualitative examples of planning behavior that complement the aggregate success rates and terminal gap measurements.

\section{Conclusion}
We present \ours, a latent world model that organizes representations into two jointly learned subspaces with linear and nonlinear dynamics. An action-conditioned linear predictor and a history-conditioned nonlinear predictor model their respective temporal evolution, while subspace-specific regularization maintains representation diversity and reduces redundancy in the linear coordinates. Experiments on four image-goal control environments demonstrate higher average planning success than the evaluated baselines under both short-range and distant-goal protocols. Ablation studies support the contributions of action conditioning and subspace-specific regularization. These findings support structuring latent representations through complementary predictive dynamics for visual planning. Our evaluation is currently limited to simulated environments. Future work will examine the generalization of \ours to real-world robotic tasks and more diverse visual conditions in tasks.

\section*{LLM Usage Statement}
During the preparation of this work, the authors used DeepSeek for language refinement and assistance with code implementation. After using this tool, the authors re-edited the whole manuscript and checked through the coding script as needed, and take full responsibility for  this article.

\appendix
\raggedbottom
\renewcommand{\thetable}{S\arabic{table}}
\renewcommand{\theequation}{S\arabic{equation}}
\renewcommand{\thefigure}{S\arabic{figure}}

\section{Proofs for the Theoretical Analysis}
\label{app:theory-proofs}
~\cref{sec:theoretical-analysis} studies the one-step latent prediction loss for the joint classes of the encoder and predictor in JEPA.
The aim is to isolate the approximation and complexity trade-off associated with the structured latent transition to an action-conditioned linear map.
This section provides detailed proofs of the theoretical results
presented in ~\cref{sec:theoretical-analysis} of the main paper. We present
the complete derivations and intermediate steps to further
clarify the theoretical results.

\subsection{Theorem 1: One-Step Generalization}

Let $S=(X_1,\ldots,X_N)$ denote a set of $N$ data sequences, where each $X_i$ represents an individual training sample.
We assume that these sequences are independently and identically distributed according to $P_X$, i.e., $S\sim P_X^N$.

Let $\mathcal J_s$ and $\mathcal J_u$ be nonempty joint classes of encoder--predictor pairs $e=(g,f)$ for the structured model and the unstructured comparator, respectively, specified before sampling.
Both classes use the same encoder family, latent split, and predictor family for the nonlinear coordinates.
We assume that the comparator can reproduce every admissible structured update with the encoder and nonlinear-coordinate predictor held fixed, giving $\mathcal J_s\subseteq\mathcal J_u$.
Although the encoder and predictor are learned from \(S\), the uniform deviation bound applies to all encoder--predictor pairs in the prescribed joint classes.

For a sequence $X$, let $\ell(e;X)$ denote the one-step latent prediction loss used in ~\cref{eq:pred}.
For $q\in\{s,u\}$ and $e\in\mathcal J_q$, define the population and empirical risks by
\[
\begin{aligned}
\mathcal R_q(e)
&=\mathbb E_{X\sim P_X}[\ell(e;X)],\\
\widehat{\mathcal R}_S(e)
&=\frac{1}{N}\sum_{i=1}^N\ell(e;X_i).
\end{aligned}
\]
We assume a common deterministic constant $M_\ell<\infty$ such that $0\leq\ell(e;X)\leq M_\ell$ for every $e\in\mathcal J_s\cup\mathcal J_u$ and every $X$ in the support of $P_X$.
This uniform bound is imposed on the admissible model classes, including their behavior on unseen sequences.

For Theorem~1, assume that the empirical risk attains its minimum over $\mathcal J_q$, and let $e_q^\star$ denote a corresponding minimizer.
Define the one-sided empirical Rademacher complexity~\cite{bartlett2002rademacher} of the loss class by
\begin{equation}
\label{eq:supp-theory-rademacher}
\widehat{\mathfrak R}_q
=\frac1N\mathbb E_{\bm\sigma}
  \sup_{e\in\mathcal J_q}\sum_{i=1}^N\sigma_i\ell(e;X_i),
\end{equation}
For fixed samples in $S$, draw $\sigma_1,\ldots,\sigma_N$ independently
from $\{-1,+1\}$, assigning probability $1/2$ to each value.
The expectation $\mathbb E_{\bm\sigma}$ averages over
these signs.

\noindent\textbf{Step 1: Symmetrization.}
For a given sample $S$, we measure the worst-case deviation between
population and empirical prediction risk over $\mathcal J_q$ in each
of the two directions. For $b\in\{-1,+1\}$, define
\[
F_q^b(S)=\sup_{e\in\mathcal J_q}
b\bigl(\mathcal R_q(e)-\widehat{\mathcal R}_S(e)\bigr).
\]
The choice $b=+1$ measures how much the empirical risk can underestimate
the population risk, and $b=-1$ measures how much it can overestimate it.
Bounding both quantities controls the absolute deviation uniformly over
$\mathcal J_q$.
In order to  bound $\mathbb E_S F_q^b(S)$, 
an independent sample
$S'=(X_i')_{i=1}^N\sim P_X^N$ is introduced. For each fixed $e\in\mathcal J_q$,
$\mathcal R_q(e)=\mathbb E_{S'}[\widehat{\mathcal R}_{S'}(e)]$.
Let $D_i(e)=\ell(e;X_i')-\ell(e;X_i)$ denote the difference between
the losses of the same model $e$ on the $i$th sequences of $S'$ and
$S$. With $S$ fixed, the average of these differences has expectation
$\mathcal R_q(e)-\widehat{\mathcal R}_S(e)$ over $S'$.
Since the supremum of an expectation is bounded
by the expectation of the supremum, we obtain
\begin{align*}
F_q^b(S)
&=\frac1N\sup_{e\in\mathcal J_q}\mathbb E_{S'}\!\left[\sum_i bD_i(e)\right]\\
&\leq\frac1N\mathbb E_{S'}\!\left[\sup_{e\in\mathcal J_q}\sum_i bD_i(e)\right].
\end{align*}
Since the sample pairs $(X_i,X_i')$ are i.i.d. and exchangeable,
their joint distribution is invariant under pairwise swaps.
Since $b\bm\sigma$ and $\bm\sigma$ have the same distribution, we obtain
\begin{align*}
\mathbb E_S F_q^b(S)
&\leq \frac1N\mathbb E_{S,S'}\sup_{e\in\mathcal J_q}\sum_i bD_i(e)\\
&= \frac1N\mathbb E_{S,S',\bm\sigma}\sup_{e\in\mathcal J_q}\sum_i \sigma_iD_i(e)\\
&\leq \frac1N\mathbb E_{S',\bm\sigma}\sup_{e\in\mathcal J_q}
        \sum_i \sigma_i\ell(e;X_i')\\
&\quad+\frac1N\mathbb E_{S,\bm\sigma}\sup_{e\in\mathcal J_q}
        \sum_i(-\sigma_i)\ell(e;X_i)\\
&=2\mathbb E_S\widehat{\mathfrak R}_q.
\end{align*}
The last inequality bounds the supremum of the sum by the sum of the
two suprema. Each term equals $\mathbb E_S\widehat{\mathfrak R}_q$.
Each term equals $\mathbb E_S\widehat{\mathfrak R}_q$, since $S$ and $S'$ have the same distribution, and the Rademacher signs are symmetric about zero.

\noindent\textbf{Step 2: Uniform deviation bound.}
Define $G_q^b(S)=F_q^b(S)-2\widehat{\mathfrak R}_q(S)$.
By the inequality above, $\mathbb E_S G_q^b(S)\leq 0$.
Replacing one sequence changes only one loss term. Since losses lie
in $[0,M_\ell]$ and $|\sigma_i|=1$, both the empirical risk and the
signed loss average change by at most $M_\ell/N$ for every model
and sign vector. Taking the supremum over $e\in\mathcal J_q$
in both $F_q^b(S)$ and $\widehat{\mathfrak R}_q(S)$,
and averaging over Rademacher signs in the latter,
preserves this bound. Thus, $G_q^b$
changes by at most $3M_\ell/N$. McDiarmid's inequality~\cite{mcdiarmid1989bounded} gives
\[
\Pr_S\bigl(G_q^b(S)>t\bigr)
\leq \exp\!\left(-\frac{2Nt^2}{9M_\ell^2}\right),
\qquad t>0.
\]
For $\delta\in(0,1)$, set
$t_\delta=3M_\ell\sqrt{\log(8/\delta)/(2N)}$. 
For each fixed $(q,b)$, a random sample $S\sim P_X^N$
satisfies $F_q^b(S)\leq 2\widehat{\mathfrak R}_q(S)+t_\delta$
with probability at least $1-\delta/8$.
A union bound over the two classes and two signs shows that all
four bounds hold on the same sample $S$ with probability at least
$1-\delta/2\geq 1-\delta$.  Hence, for each
$q\in\{s,u\}$ and every $e\in\mathcal J_q$, we have
\[
\bigl|\mathcal R_q(e)-\widehat{\mathcal R}_S(e)\bigr|
\leq 2\widehat{\mathfrak R}_q(S)+t_\delta.
\]

\noindent\textbf{Step 3: Prediction risk bound.}
Denote $\mathcal R_q^\star=\inf_{e\in\mathcal J_q}\mathcal R_q(e)$
and  $\xi_q=\widehat{\mathcal R}_S(\widehat e_q)
-\inf_{e\in\mathcal J_q}\widehat{\mathcal R}_S(e)$ be the empirical
prediction excess of any fit $\widehat e_q\in\mathcal J_q$.
By the preceding inequality, we have
\begin{equation}
\label{eq:supp-theory-risk}
\mathcal R_q(\widehat e_q)\leq
\mathcal R_q^\star+\xi_q+4\widehat{\mathfrak R}_q+2t_\delta.
\end{equation}
\cref{eq:supp-theory-risk} applies to any fitted pair $\widehat e_q\in\mathcal J_q$, including a model obtained by regularized training.
The term $\xi_q$ measures empirical prediction excess relative to the best prediction fit in $\mathcal J_q$ and may reflect both representation regularization and incomplete optimization.
Theorem~1 specializes this bound to the idealized reference solution with $\xi_q=0$.

\subsection{Lemma 2: Structured Loss Complexity Bound}
\noindent\textbf{Loss decomposition.}
Given with dataset $S$, for $e\in\mathcal J_s$, let $\ell_i^{\mathrm{lin}}(e)$
and $\ell_i^{\mathrm{nlin}}(e)$ denote the one-step  prediction errors
of the linear and nonlinear predictions in Eqs.~\eqref{eq:linear} and~\eqref{eq:nonlinear}, evaluated
on $X_i$. The full prediction concatenates the two outputs, such that
\[
\ell(e;X_i)=\ell_i^{\mathrm{lin}}(e)+\ell_i^{\mathrm{nlin}}(e).
\]
We denote $\widehat{\mathfrak R}_{\mathrm{lin}}$ and
$\widehat{\mathfrak R}_{\mathrm{nlin}}$ as the empirical Rademacher
complexities in Eq. \eqref{eq:supp-theory-rademacher} with the loss replaced
by the corresponding branch loss over $\mathcal J_s$. Bounding the
supremum of the sum by the sum of the two suprema gives
\[
\widehat{\mathfrak R}_s
\leq \widehat{\mathfrak R}_{\mathrm{lin}}
    +\widehat{\mathfrak R}_{\mathrm{nlin}}.
\]
To bound $\widehat{\mathfrak R}_{\mathrm{lin}}$, we first bound the
dimension of the loss span for a fixed encoder. As the encoder varies,
we approximate the  projectors using a finite set of
representative projectors.

\noindent\textbf{Loss subspaces and a finite cover of projectors.}
Let $\mathcal G$ be the set of encoders in $\mathcal J_s$. For each
$g\in\mathcal G$, 
the loss vector
$\bm\ell^{\mathrm{lin}}(e)=(\ell_i^{\mathrm{lin}}(e))_{i=1}^N$
records the linear prediction losses on $S$ as the predictor varies
with $g$ fixed. Subtracting $(M_\ell/2)\bm 1$ from each loss vector
bounds each shifted entry in magnitude by $M_\ell/2$ and preserves the empirical Rademacher complexity, since it adds the same mean-zero term to every signed loss sum. Define
\[
V_g=\operatorname{span}\bigl(
\{\bm 1\}\cup
\{\bm\ell^{\mathrm{lin}}(e):e=(g,f)\in\mathcal J_s\}\bigr).
\]
Let $P_g$ be the orthogonal projector onto $V_g$, and let
$\rho_{\mathrm{lin}}=\max_{g\in\mathcal G}\dim V_g$.
For $\varepsilon>0$, let
$\mathcal P_\varepsilon\subseteq\{P_g:g\in\mathcal G\}$ be a smallest
set such that every $P_g$ is within operator-norm distance
$\varepsilon$ of some $P\in\mathcal P_\varepsilon$.
Define the covering number
$C_{\mathrm{lin}}(\varepsilon)=|\mathcal P_\varepsilon|$ as the minimum
number of representative projectors.
Both $\rho_{\mathrm{lin}}$ and $C_{\mathrm{lin}}(\varepsilon)$ depend on the realized sample $S$ and the full admissible encoder family $\mathcal G$.
Fixing an encoder is an intermediate step in the construction, and the final cover ranges over $\{P_g:g\in\mathcal G\}$.
The covering term therefore accounts for variation across encoders, including the encoder selected using $S$.

\noindent\textbf{Step 1: Dimension bound for the loss subspaces.}
Given with a fixed
encoder $g$, we firstly bound $\dim V_g$ . Let $\bm z_i^{\mathrm{lin}}$ and
$\bm z_{i,+}^{\mathrm{lin}}$ be the current and next linear latents in
$X_i$, and let $\bm a_i$ be its current action. By ~\cref{eq:linear}, the
linear prediction residual is $W\bm v_i$, where
\begin{align*}
\bm v_i &=
\bigl[[1;\bm a_i]\otimes\bm z_i^{\mathrm{lin}};
                  \bm z_{i,+}^{\mathrm{lin}}\bigr],\\
W &=[K_0\ \cdots\ K_m\ {-I_{d_{\mathrm{lin}}}}],
\end{align*}
such that  $\ell_i^{\mathrm{lin}}(e)=\bm v_i^\top W^\top W\bm v_i$
The vectors $\bm v_i$ are fixed once $g$ and $S$ are given.
Writing $p=(m+2)d_{\mathrm{lin}}$, all linear loss vectors therefore
lie in the image of the linear map
$A\mapsto(\bm v_i^\top A\bm v_i)_{i=1}^N$
on symmetric $p\times p$ matrices. That space of matrices has
dimension $p(p+1)/2$. 
So $\rho_{\mathrm{lin}}\leq\min\{N,1+p(p+1)/2\}$.

\noindent\textbf{Step 2: Reduction to a finite cover.}
For a given sign vector $\bm\sigma$, let
$Z=\max_{P\in\mathcal P_\varepsilon}\|P\bm\sigma\|_2$ be the
largest norm obtained by projecting it with the representatives
in $\mathcal P_\varepsilon$. For each $g$, choose a representative
$P$ with $\|P_g-P\|_{\mathrm{op}}\leq\varepsilon$. Since
$\|\bm\sigma\|_2=\sqrt N$, the triangle inequality gives
\[
\|P_g\bm\sigma\|_2
\leq \|P\bm\sigma\|_2+\|(P_g-P)\bm\sigma\|_2
\leq Z+\varepsilon\sqrt N.
\]
The shifted loss vector
$\bm\ell_c^{\mathrm{lin}}(e)=\bm\ell^{\mathrm{lin}}(e)-(M_\ell/2)\bm 1$
has norm at most $M_\ell\sqrt N/2$ because each entry has magnitude
at most $M_\ell/2$. It lies in $V_g$, while
$\bm\sigma-P_g\bm\sigma$ is orthogonal to $V_g$. Thus
$\langle\bm\ell_c^{\mathrm{lin}}(e),\bm\sigma\rangle
=\langle\bm\ell_c^{\mathrm{lin}}(e),P_g\bm\sigma\rangle.$
By Cauchy--Schwarz, this inner product is at most
$(M_\ell\sqrt N/2)(Z+\varepsilon\sqrt N)$ for every $e$.
Applying Eq. \eqref{eq:supp-theory-rademacher} to the preceding bound on the centered inner product gives
\[
\widehat{\mathfrak R}_{\mathrm{lin}}
\leq \frac{M_\ell}{2}
\left(\varepsilon+\frac{\mathbb E_{\bm\sigma}Z}{\sqrt N}\right).
\]

\noindent\textbf{Step 3: Bound on the expected maximum projected norm.}
For $P\in\mathcal P_\varepsilon$ and $0<\alpha<1/2$, let
$\bm w\sim\mathcal N(0,I_N)$ be independent of $\bm\sigma$ and set
$r=\operatorname{rank}(P)\leq\rho_{\mathrm{lin}}$. Applying the
Gaussian moment-generating function and averaging over the
independent signs gives
\begin{align*}
\mathbb E_{\bm\sigma} e^{\alpha\|P\bm\sigma\|_2^2}
&=\mathbb E_{\bm w}\prod_i
  \cosh\bigl(\sqrt{2\alpha}(P\bm w)_i\bigr)\\
&\leq \mathbb E_{\bm w}e^{\alpha\|P\bm w\|_2^2}\\
&=(1-2\alpha)^{-r/2}
\leq (1-2\alpha)^{-\rho_{\mathrm{lin}}/2}.
\end{align*}
Here we use $\cosh v\leq e^{v^2/2}$ and
$\|P\bm w\|_2^2\sim\chi_r^2$, where $\chi_r^2$ denotes the
chi-square distribution with $r$ degrees of freedom.
Bounding $e^{\alpha Z^2}$ by the sum of these exponentials over the
representatives and applying Jensen's inequality gives
\begin{align*}
(\mathbb E_{\bm\sigma}Z)^2
&\leq\frac{\log C_{\mathrm{lin}}(\varepsilon)
-(\rho_{\mathrm{lin}}/2)\log(1-2\alpha)}{\alpha}\\
&\leq\frac{\log C_{\mathrm{lin}}(\varepsilon)}{\alpha}
+\frac{\rho_{\mathrm{lin}}}{1-2\alpha}.
\end{align*}
The bound holds for every $0<\alpha<1/2$, while $Z$ is independent
of $\alpha$. Setting $t=2\alpha$ puts the two terms in the form
$a/t+b/(1-t)$, with $a=2\log C_{\mathrm{lin}}(\varepsilon)$ and
$b=\rho_{\mathrm{lin}}$. Cauchy--Schwarz gives the lower bound
$(\sqrt a+\sqrt b)^2$, which is reached in the limit
$t\to\sqrt a/(\sqrt a+\sqrt b)$. Taking the square root of this
infimum gives
\[
\mathbb E_{\bm\sigma}Z
\leq \sqrt{\rho_{\mathrm{lin}}}
     +\sqrt{2\log C_{\mathrm{lin}}(\varepsilon)}.
\]
Combining this estimate with the linear complexity bound and the
loss decomposition yields Eq.~\eqref{eq:theory-split-complexity} of Lemma~2,
\[
\widehat{\mathfrak R}_s
\leq \widehat{\mathfrak R}_{\mathrm{nlin}}
     +B_{d_{\mathrm{lin}}}(\varepsilon),
\]
where
$B_{d_{\mathrm{lin}}}(\varepsilon)
=\frac{M_\ell}{2}\left(
\varepsilon+
\frac{\sqrt{\rho_{\mathrm{lin}}}+
      \sqrt{2\log C_{\mathrm{lin}}(\varepsilon)}}{\sqrt N}
\right).$

\subsection{Corollary 3: Upper-Bound Comparison}
\label{app:corollary}
By Eq. \eqref{eq:supp-theory-rademacher} with $q=u$,
for each sign vector $\bm\sigma$ we take the supremum of
$N^{-1}\sum_{i=1}^N \sigma_i\ell(e;X_i)$ over $e\in\mathcal J_u$
and then average over $\bm\sigma$. To lower-bound
$\widehat{\mathfrak R}_u$, we restrict this supremum to a subset of
comparator models with a fixed encoder $g\in\mathcal G$.
For such a model $e=(g,f)$, decompose $\ell(e;X_i)$ as in Lemma~2
into the  prediction errors $\ell_i^{\mathrm{lin}}(e)$ and
$\ell_i^{\mathrm{nlin}}(e)$ on the two sets of latent coordinates.
The superscript $\mathrm{lin}$ refers to the coordinates predicted
by the linear update in Eq.~\eqref{eq:linear} in the structured model. The
comparator uses a nonlinear predictor for the same coordinates.

At the fixed encoder $g$, we make the following additional assumptions for the upper-bound comparison.

\noindent \emph{Simultaneous realizability.}
Let $I\subseteq\{1,\ldots,N\}$ contain $k$ selected sequences.
Assume that there are prescribed loss levels $0\leq\ell_i^-<\ell_i^+\leq M_\ell$ with $\ell_i^+-\ell_i^-\geq\beta>0$ for every $i\in I$.
For every assignment $b\in\{-,+\}^{I}$, assume that a comparator predictor $h_b$ for the coordinates denoted by $z^{\mathrm{lin}}$ can attain the prediction loss $\ell_i^{b_i}$ at every selected sequence simultaneously.
This requires one predictor for each complete assignment, rather than separate predictors chosen independently for individual sequences.

\noindent \emph{Independent branch pairing.}
Assume further that one such $h_b$ can be chosen for each assignment so that pairing it with any nonlinear-coordinate predictor  at the same encoder $g$ yields an encoder--predictor pair in $\mathcal J_u$.
Each component retains its input--output map under this pairing, and the combined model obeys the same uniform loss bound.
We fix the selected sequences, the prescribed loss levels, and one such predictor for each assignment before drawing the Rademacher signs $\sigma$.

\noindent\textbf{Step 1: Comparator complexity lower bound.}
Given $\bm\sigma_I=(\sigma_i)_{i\in I}$, use the predictor fixed
for the loss combination prescribed by these signs, and let
$u_i(\bm\sigma_I)$ be its loss on the linear coordinates of $X_i$
for each $i=1,\ldots,N$. For $i\in I$, the signed loss is
$\ell_i^+$ or $-\ell_i^-$ with equal probability. For $i\notin I$,
$\sigma_i$ has mean zero and is independent of $u_i(\bm\sigma_I)$,
which depends only on the signs in $I$. Thus
\[
\mathbb E_{\bm\sigma}[\sigma_i u_i(\bm\sigma_I)]
=
\begin{cases}
(\ell_i^+-\ell_i^-)/2, & i\in I,\\
0, & i\notin I.
\end{cases}
\]

By the pairing assumption, for each $\bm\sigma$ we can lower-bound
the supremum over $\mathcal J_u$ by holding the chosen predictor for
$\bm z^{\mathrm{lin}}$ fixed and taking the supremum over the
structured predictors for $\bm z^{\mathrm{nlin}}$ at $g$.
The term $\sum_i\sigma_i u_i(\bm\sigma_I)$ is constant over this
restricted set and can be taken outside the supremum. Let
$\widehat{\mathfrak R}_{\mathrm{nlin}}(g)$ be the complexity from
Lemma~2 restricted to encoder $g$. Averaging over the signs and
using the loss gap on each of the $k$ selected sequences gives
\begin{equation}
\label{eq:supp-theory-comparator-lower}
\begin{aligned}
\widehat{\mathfrak R}_u
&\geq \widehat{\mathfrak R}_{\mathrm{nlin}}(g)
+\frac1N\mathbb E_{\bm\sigma}\sum_{i=1}^N\sigma_i u_i(\bm\sigma_I)\\
&\geq \widehat{\mathfrak R}_{\mathrm{nlin}}(g)+\frac{k\beta}{2N}.
\end{aligned}
\end{equation}

\noindent\textbf{Step 2: Upper-bound comparison.}
For the empirical prediction minimizers in Theorem~1, let $U_u$
be the right-hand side of Eq.~\eqref{eq:theory-risk-bound} for the comparator and let
$U_s(\varepsilon)$ be the corresponding bound for the structured
model after substituting Eq.~\eqref{eq:theory-split-complexity} for its complexity term.
Both bounds hold on the same event from Theorem~1. Their confidence
terms cancel upon subtraction because $N$, $M_\ell$, and $\delta$
are shared. Let
$\Delta_{\mathrm{app}}=\mathcal R_s^\star-\mathcal R_u^\star$.
By Eq. \eqref{eq:supp-theory-comparator-lower}, we have
\begin{align*}
\frac{U_u-U_s(\varepsilon)}{4}
&= \widehat{\mathfrak R}_u-\widehat{\mathfrak R}_{\mathrm{nlin}} -B_{d_{\mathrm{lin}}}(\varepsilon)
  -\frac{\Delta_{\mathrm{app}}}{4}\\[2pt]
&\geq \frac{k\beta}{2N}
  -\bigl(\widehat{\mathfrak R}_{\mathrm{nlin}}
         -\widehat{\mathfrak R}_{\mathrm{nlin}}(g)\bigr)\\
&\quad -B_{d_{\mathrm{lin}}}(\varepsilon)
  -\frac{\Delta_{\mathrm{app}}}{4}.
\end{align*}
The 
term in the lower bound that varies with $\varepsilon$ is
$-B_{d_{\mathrm{lin}}}(\varepsilon)$. To obtain the supremum of
the lower bound over $\varepsilon>0$, we replace
$B_{d_{\mathrm{lin}}}(\varepsilon)$ by its infimum. The resulting
expression is strictly positive precisely when Eq.~\eqref{eq:corollary} holds, such that
\begin{align*}
\frac{k\beta}{2N}
> \inf_{\varepsilon>0} B_{d_{\mathrm{lin}}}(\varepsilon)
+\frac{\Delta_{\mathrm{app}}}{4}
  +\bigl(\widehat{\mathfrak R}_{\mathrm{nlin}}
         -\widehat{\mathfrak R}_{\mathrm{nlin}}(g)\bigr).
\end{align*}
Thus, under Eq.~\eqref{eq:corollary}, the lower bound is strictly positive for some
$\varepsilon>0$, yielding $U_s(\varepsilon)<U_u$.

\noindent\textbf{Interpretation of Corollary~3.}
The argument gives a sufficient condition for ordering the explicit upper bounds $U_s(\varepsilon)$ and $U_u$ for the 
empirical prediction minimizers.
 The left-hand side of Eq.~\eqref{eq:corollary}, $k\beta/(2N)$, lower-bounds the
complexity contribution of the comparator's predictor on the linear
coordinates. With the encoder fixed, the
comparator can realize every combination of high and low losses across $k$
sampled sequences, with a separation of at least $\beta$ on each sequence.
On the right-hand side, $\Delta_{\mathrm{app}}/4$ is the increase in the
best achievable population risk of the joint encoder--predictor $e=(g,f)$ class,
scaled by the coefficient of the complexity term in Eq.~\eqref{eq:theory-risk-bound}. This increase
can be small when the encoder learns coordinates that are well predicted by
the linear update and the nonlinear branch can model the remaining
coordinates. The term $\inf_{\varepsilon>0} B_{d_{\mathrm{lin}}}(\varepsilon)$
bounds the complexity of the structured linear branch. For a fixed
covering radius $\varepsilon$, $B_{d_{\mathrm{lin}}}(\varepsilon)$ is
smaller when the maximum dimension $\rho_{\mathrm{lin}}$ of the loss
spans at fixed encoders is small and when the projectors onto these spans
across encoders can be approximated by few representatives, yielding a
small covering number $C_{\mathrm{lin}}(\varepsilon)$. Finally,
$\widehat{\mathfrak R}_{\mathrm{nlin}}-\widehat{\mathfrak R}_{\mathrm{nlin}}(g)$
accounts for the additional nonlinear-loss complexity allowed by varying
the encoder. The structured bound is tighter when the comparator's
contribution $k\beta/(2N)$ exceeds the sum of these three terms.

\section{Detailed Setups and Additional Results}
\label{app:implementation-details}

\subsection{Implementation Details}

\paragraph{Training and Planning Protocol.}
The training and planning protocols of this work mainly follow LeWM \cite{maes2026lewm}, which is a strong baseline in the field compared by many recent work \cite{thil2026sdjepa}, where we adopt ts dataset preparation procedures, unless otherwise specified. We consider two planning settings based on the temporal distance between the start and goal states along demonstration trajectories: short-range, with a distance of 25 steps, and distant-goal, with a distance of 100 steps.  The short-range setting follows LeWM, while the distant-goal evaluation datasets are constructed as~\cite{zhao2026singlelatentspaceduallatent}.

\paragraph{Hyperparameter Settings.}
We apply a frame-skip of 5 and train \ours for 10 epochs with history length $H=3$ model steps using three random seeds \(0,42,3072\) and evaluate  the final checkpoint of each run using a fixed planning seed of seed 42.  Training uses the AdamW~\cite{loshchilov2019decoupled} optimizer with a batch size of 128 and a weight decay of $10^{-3}$.
We partition the 192-dimensional latent representation  into linear and nonlinear subspaces. We select the linear subspace dimension for each environment based on the sensitivity analysis in Fig.~2 of the main paper. We use $\dimref=24$ for PushT and $\dimref=48$ for TwoRoom, Reacher, and Cube. With the total latent dimension fixed at 192, the corresponding nonlinear subspace dimensions are 168 and 144, respectively. We compute SIGReg using 1,024 random projections and assign it a loss weight of $\lambda=0.09$. The variance and correlation losses are weighted by $\lambda_{\rm var}=0.10$ and $\lambda_{\rm corr}=0.01$, respectively.
For planning, we take the CEM configuration used in LeWM \cite{maes2026lewm}. At each planning step, we optimize an action sequences over horizon of $H_{p}=5$ model steps, sampling 300 candidates per CEM iteration. The main modification lies in the number of optimization iterations: we allow up to 30 iterations for all methods reported in the main results across all four environments, whereas the setting in LeWM uses 30 iterations for PushT and 10 for the other three environments.

\begin{figure*}[t]
    \centering
    \includegraphics[width=0.8\linewidth]{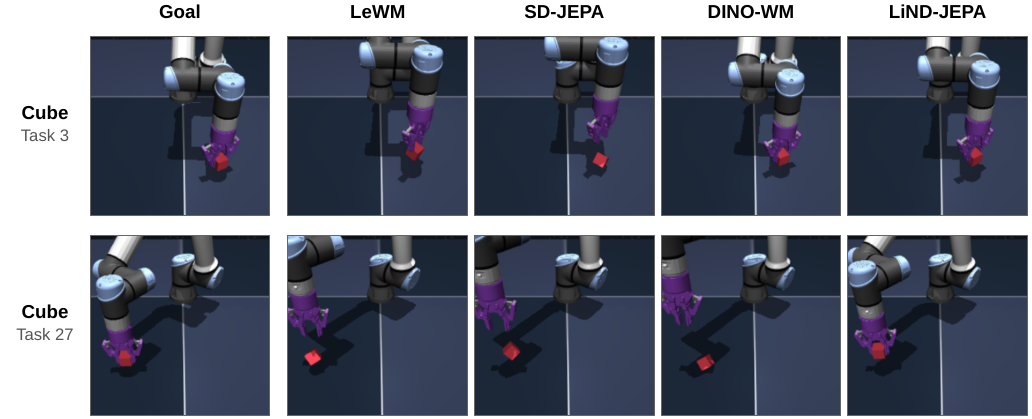}
    \caption{
    \textbf{``Successful''  planning in the Cube environment.} In the presented tasks, the baseline
    rollouts satisfy the original cube-position criterion with the cube
    ungrasped, even though the goal requires a grasp. \ours reaches
    the target while holding the cube, matching the goal grasp state.}
    \label{fig:cube-gripper}
\end{figure*}

\paragraph{Baseline Implementations.}
For LeWM \cite{maes2026lewm} and SD-JEPA \cite{thil2026sdjepa}, we use their official implementations. For DINO-WM \cite{zhou2025dinowm}, we adapt the authors' implementation by removing proprioceptive inputs while retaining the frozen, pretrained DINOv2 ViT-S/14 encoder. To ensure consistent evaluation, we apply the same data preprocessing and key planning configurations across all baselines and \ours and align training settings where feasible, while preserving method-specific objectives and planning costs.

\paragraph{Computing Environment.}
\ours  are conducted using NVIDIA A100-SXM4 GPUs with 80\,GB of memory, with each training run using a single GPU. The experiments are implemented using Python 3.12.14, PyTorch 2.11.0 built upon CUDA 12.8, and torchvision 0.26.0.

\subsection{Additional Results and Further Discussions}
\label{app:additional-results}

\paragraph{Grasp-Aware Goal Evaluation in Cube.}
In Cube experiments, the  success criterion considered a rollout successful once the cube reached its target position, regardless of the arm configuration or gripper state. For example, in task~3 and task~27 of Fig.~\ref{fig:cube-gripper}, several rollouts left the cube near its target with the gripper disengaged, although the goal frame shows the cube held between the fingers. 
The original criterion thus fails to fully capture the goal image's requirement that the gripper continuously hold the cube while transporting it to the target position.
To better reflect the target state depicted in the goal image, we revise the success criterion by retaining the original cube-position requirement and adding three constraints on the gripper’s position, opening angle, and grasp state.
Specifically, a rollout is considered successful only if
all four conditions are simultaneously satisfied at an
executed step within the rollout budget:
\textit{(i)} the cube center is within $4\,\mathrm{cm}$ of its goal position (original criterion);
\textit{(ii)} the gripper's pinch center is within $3\,\mathrm{cm}$ of its position in the goal frame;
\textit{(iii)} the normalized gripper opening angle differs from the goal by at most $0.10$; and \textit{(iv)} the grasp state matches that of the goal frame.
A grasp is considered valid when both finger pads contact the cube, each with a normal force exceeding $0.001\,\mathrm{N}$.
We rerun evaluation with the same model checkpoints, planner settings, initial states, and goal images, and score each rollout under the grasp-aware criterion.
{As shown in Tab.~\ref{tab:cube-gripper-results}, \ours exceeds the strongest baseline by $29.33$\%
at short-range and $14.00$\%
at distant-goal. These gains further support the potential of structured  dynamics in downstream planning.
}

\begin{table}[t]
    \centering
    \small
    \setlength{\tabcolsep}{4pt}
    \renewcommand{\arraystretch}{1.12}
    \caption{Cube planning success rates (\%) under the original
    and revised grasp-aware criteria. The original-criterion results  are taken from Tabs.~\ref{tab:planning-offset25} and~\ref{tab:planning-offset100} of the main paper.}
    \label{tab:cube-gripper-results}
    \begin{tabular*}{0.9\columnwidth}{
        @{\hspace{5pt}}
        l@{\extracolsep{\fill}}cc
        @{\hspace{5pt}}
    }
        \toprule
        Method & Short-range & Distant-goal \\
        \midrule
        \multicolumn{3}{@{\hspace{5pt}}l}{\textit{Original criterion}} \\
        Random policy & $43.33 \pm 1.15$ & $36.00 \pm 2.00$ \\
        LeWM          & $68.00 \pm 5.29$ & $58.00 \pm 2.00$ \\
        SD-JEPA       & $68.00 \pm 5.29$ & $62.67 \pm 5.03$ \\
        DINO-WM       & $76.00 \pm 5.29$ & $64.67 \pm 6.43$ \\
        LiND-JEPA     & {\boldmath$92.00 \pm 3.46$}
                      & {\boldmath$72.67 \pm 1.15$} \\
        \midrule
        \multicolumn{3}{@{\hspace{5pt}}l}{\textit{Revised criterion}} \\
        Random policy & $0.67 \pm 1.15$  & $2.00 \pm 0.00$  \\
        LeWM          & $34.67 \pm 3.06$ & $36.00 \pm 2.00$ \\
        SD-JEPA       & $31.33 \pm 4.16$ & $40.67 \pm 11.02$ \\
        DINO-WM       & $22.00 \pm 4.00$     & $30.00 \pm 3.46$   \\
        LiND-JEPA     & {\boldmath$64.00 \pm 5.29$}
                      & {\boldmath$54.67 \pm 2.31$} \\
        \bottomrule
    \end{tabular*}
\end{table}

\begin{figure*}[t]
    \centering
    \includegraphics[width=0.9\linewidth]{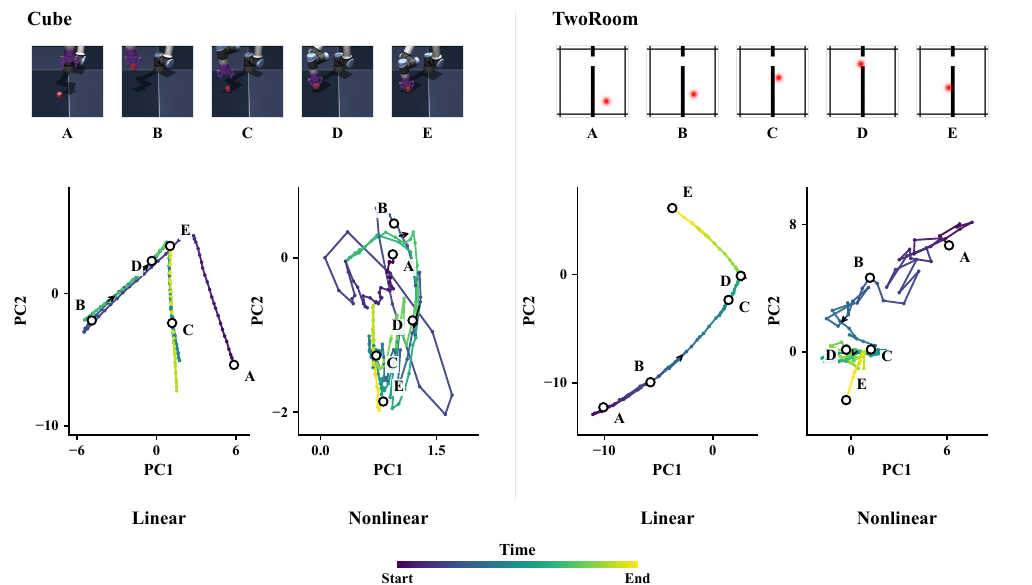}
    \caption{\textbf{PCA projections of the linear and nonlinear latent components in Cube (left) and TwoRoom (right).} Color indicates episode progress, and markers A--E link the five observations to their positions along both trajectories.}
    \label{fig:latent-trajectories}
\end{figure*}

\paragraph{Visualization of Latent Trajectories.}
To illustrate how the learned representations $\zref$ and $\zcont$ evolve across different tasks, we visualize their trajectories in Cube, a 3D object manipulation task, and TwoRoom, a 2D navigation task. We use the frozen \ours encoder to extract both representations from observed frames and visualize their trajectories in a two-dimensional space. Specifically, We fit a separate two-dimensional PCA projection to each component using episodes~0--19 and applied the resulting projections to every frame of episode~20. As shown in  Fig.~\ref{fig:latent-trajectories},  in Cube,  the projected trajectory of $\zref$ exhibits several nearly straight segments with  changes in direction. In TwoRoom, it bends as the agent reaches the doorway at marker~D. These patterns are consistent with the intended role of the structured predictor, which encourages the encoder to learn a representation $\zref$ whose transitions can be modeled by an action-conditioned linear map.

\begin{figure}[t]
    \centering
    \includegraphics[width=0.95\linewidth]{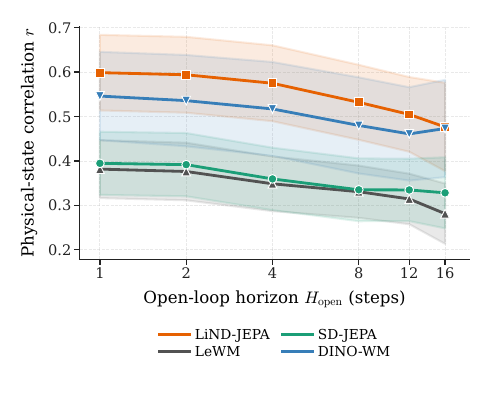}
    \caption{\textbf{Open-loop readout of physical-state.}
    Each point averages four environments and three training seeds. The band is the standard error across these twelve runs.
    We fit a linear probe on the first 24 principal components of the encoded training latents and use the same PCA projection and probe for open-loop predictions.
    The targets here are agent position, finger offset, or block position and yaw.}
    \label{fig:state-fidelity}
\end{figure}

\paragraph{Physical-State Correlation in Open-Loop Rollouts.}
\cref{fig:state-fidelity} uses a linear probe to measure how the correlation between the predicted latent $\widehat{\bm z}$ and the ground-truth physical coordinates changes during open-loop rollout. We compare \ours with LeWM, SD-JEPA, and DINO-WM. Given the encoded real context, each model predicts the future latents from the recorded actions for $H_{\rm open}$ model steps and does not observe the intermediate frames. We fit a ridge probe on latent states encoded from the training episodes and report the mean Pearson correlation of prediction on held-out episodes, following~\cite{zhou2025dinowm,terver2026jepawm}.

The correlation decreases with $H_{\rm open}$ for all four models, which is consistent with the accumulation of open-loop prediction error. At most the horizons, \ours stays above all three baselines. 
The higher correlations suggest that physical coordinates remain more readily decodable from the predicted latents of \ours. The structured dynamics of the linear subspace $\zref$ may help preserve this decodability over multiple prediction steps.

\begin{table*}[t]
\centering
\caption{Physical-state probing in Cube. Across three training seeds, MSE is reported as mean $\pm$ sample standard deviation, while Pearson $r$ is reported as  mean. For each seed, overall NMSE and Pearson $r$ are averaged equally across  eight properties, with each property's NMSE computed by averaging coordinate-wise MSEs normalized by their respective training-set variances. Best results are in bold.}
\label{tab:probe_cube}
\small
\setlength{\tabcolsep}{9pt}
\begin{tabular}{@{}llcccc@{}}
\toprule
 & & \multicolumn{2}{c}{Linear} & \multicolumn{2}{c}{MLP} \\
\cmidrule(lr){3-4}\cmidrule(lr){5-6}
Property & Representation & MSE $\downarrow$ & $r\uparrow$ &
MSE $\downarrow$ & $r\uparrow$ \\
\midrule
Joint position  & LeWM & $4.904\pm 0.019$ & $0.70096$ & $4.978\pm 0.001$ & $0.70584$ \\
 & Ours ($\zref$) & $4.846\pm 0.011$ & $0.69813$ & $4.850\pm 0.030$ & $0.71141$ \\
 & Ours ($\zcont$) & $4.527\pm 0.015$ & $0.71303$ & $4.739\pm 0.084$ & $\mathbf{0.72478}$ \\
 & Ours ($\bm{z}$) & $\mathbf{4.512\pm 0.011}$ & $\mathbf{0.72404}$ & $\mathbf{4.612\pm 0.034}$ & $0.72348$ \\
\addlinespace[2pt]
Joint velocity  & LeWM & $1.407\pm 0.001$ & $0.15005$ & $1.236\pm 0.025$ & $0.33181$ \\
 & Ours ($\zref$) & $1.379\pm 0.011$ & $0.18241$ & $1.198\pm 0.032$ & $0.32444$ \\
 & Ours ($\zcont$) & $1.415\pm 0.001$ & $0.12644$ & $1.113\pm 0.042$ & $0.36037$ \\
 & Ours ($\bm{z}$) & $\mathbf{1.297\pm 0.006}$ & $\mathbf{0.25461}$ & $\mathbf{1.051\pm 0.032}$ & $\mathbf{0.39502}$ \\
\addlinespace[2pt]
End-effector position  & LeWM & $2.733\pm 0.193$ & $0.99057$ & $\mathbf{0.148\pm 0.021}$ & $0.99905$ \\
 & Ours ($\zref$) & $3.473\pm 0.317$ & $0.98780$ & $0.580\pm 0.108$ & $0.99756$ \\
 & Ours ($\zcont$) & $4.688\pm 0.310$ & $0.98267$ & $0.202\pm 0.029$ & $0.99899$ \\
 & Ours ($\bm{z}$) & $\mathbf{2.609\pm 0.249}$ & $\mathbf{0.99185}$ & $0.182\pm 0.034$ & $\mathbf{0.99910}$ \\
\addlinespace[2pt]
End-effector yaw & LeWM & $2.870\pm 0.004$ & $-0.02413$ & $2.931\pm 0.046$ & $0.00907$ \\
 & Ours ($\zref$) & $2.839\pm 0.000$ & $0.03909$ & $2.861\pm 0.004$ & $0.01174$ \\
 & Ours ($\zcont$) & $2.670\pm 0.007$ & $0.25361$ & $2.676\pm 0.045$ & $0.25735$ \\
 & Ours ($\bm{z}$) & $\mathbf{2.669\pm 0.007}$ & $\mathbf{0.25411}$ & $\mathbf{2.665\pm 0.032}$ & $\mathbf{0.26436}$ \\
\addlinespace[2pt]
Gripper opening  & LeWM & $5.871\pm 0.671$ & $0.95270$ & $1.547\pm 0.176$ & $0.98779$ \\
 & Ours ($\zref$) & $2.405\pm 0.160$ & $0.98090$ & $0.980\pm 0.042$ & $0.99230$ \\
 & Ours ($\zcont$) & $3.756\pm 0.226$ & $0.97001$ & $0.874\pm 0.051$ & $0.99312$ \\
 & Ours ($\bm{z}$) & $\mathbf{1.993\pm 0.120}$ & $\mathbf{0.98421}$ & $\mathbf{0.828\pm 0.054}$ & $\mathbf{0.99349}$ \\
\addlinespace[2pt]
Block position  & LeWM & $1.213\pm 0.182$ & $0.99517$ & $0.099\pm 0.007$ & $0.99946$ \\
 & Ours ($\zref$) & $0.399\pm 0.029$ & $0.99787$ & $0.089\pm 0.005$ & $0.99946$ \\
 & Ours ($\zcont$) & $1.071\pm 0.080$ & $0.99480$ & $0.060\pm 0.001$ & $0.99965$ \\
 & Ours ($\bm{z}$) & $\mathbf{0.239\pm 0.012}$ & $\mathbf{0.99882}$ & $\mathbf{0.053\pm 0.001}$ & $\mathbf{0.99969}$ \\
\addlinespace[2pt]
Block quaternion  & LeWM & $1.613\pm 0.001$ & $0.01175$ & $1.650\pm 0.027$ & $0.01704$ \\
 & Ours ($\zref$) & $1.618\pm 0.003$ & $0.00638$ & $\mathbf{1.616\pm 0.006}$ & $0.02801$ \\
 & Ours ($\zcont$) & $\mathbf{1.611\pm 0.002}$ & $0.02860$ & $1.639\pm 0.021$ & $0.03203$ \\
 & Ours ($\bm{z}$) & $1.614\pm 0.004$ & $\mathbf{0.03471}$ & $1.631\pm 0.015$ & $\mathbf{0.03286}$ \\
\addlinespace[2pt]
Block yaw & LeWM & $\mathbf{3.231\pm 0.015}$ & $\mathbf{-0.00771}$ & $3.324\pm 0.006$ & $0.01353$ \\
 & Ours ($\zref$) & $3.245\pm 0.008$ & $-0.02966$ & $\mathbf{3.244\pm 0.016}$ & $0.01353$ \\
 & Ours ($\zcont$) & $3.252\pm 0.004$ & $-0.03420$ & $3.346\pm 0.010$ & $0.02168$ \\
 & Ours ($\bm{z}$) & $3.261\pm 0.004$ & $-0.03633$ & $3.343\pm 0.015$ & $\mathbf{0.02277}$ \\
\addlinespace[2pt]
\midrule
\textbf{Overall (NMSE)} & LeWM & $0.671\pm 0.001$ & $0.47117$ & $0.656\pm 0.002$ & $0.50795$ \\
 & Ours ($\zref$) & $0.662\pm 0.002$ & $0.48287$ & $0.643\pm 0.003$ & $0.50981$ \\
 & Ours ($\zcont$) & $0.661\pm 0.002$ & $0.50437$ & $0.634\pm 0.003$ & $0.54850$ \\
 & Ours ($\bm{z}$) & $\mathbf{0.645\pm 0.002}$ & $\mathbf{0.52575}$ & $\mathbf{0.629\pm 0.002}$ & $\mathbf{0.55385}$ \\
\bottomrule
\end{tabular}
\end{table*}

\begin{table}[t]
\centering
\caption{
Physical-state probing  in TwoRoom. Across three training seeds, MSE is reported as mean $\pm$ sample standard deviation, while Pearson $r$ is reported as mean. Best results are in bold.
}
\label{tab:probe_tworoom}

\setlength{\tabcolsep}{4pt}
\renewcommand{\arraystretch}{1.12}

\resizebox{\columnwidth}{!}{%
\begin{tabular}{@{}lcccc@{}}
\toprule
& \multicolumn{2}{c}{Linear}
& \multicolumn{2}{c}{MLP} \\
\cmidrule(lr){2-3}
\cmidrule(l){4-5}

Representation
& MSE $\downarrow$
& $r$ $\uparrow$
& MSE $\downarrow$
& $r$ $\uparrow$ \\
\midrule

LeWM
& $5.1230 \pm 1.3800$
& $0.998296$
& $0.2285 \pm 0.0247$
& $0.999924$ \\

Ours ($\zref$)
& $1.0640 \pm 0.1060$
& $0.999643$
& $0.2312 \pm 0.0244$
& $0.999926$ \\

Ours ($\zcont$)
& $1.3210 \pm 0.2450$
& $0.999567$
& $0.1100 \pm 0.0069$
& $0.999964$ \\

Ours ($\bm{z}$)
& $\mathbf{0.2267 \pm 0.0027}$
& $\mathbf{0.999927}$
& $\mathbf{0.0996 \pm 0.0060}$
& $\mathbf{0.999968}$ \\

\bottomrule
\end{tabular}%
}
\end{table}

\paragraph{Probing.}
We probe the physical information encoded in the learned structured latent representations. We train linear and nonlinear (with MLP) probes on the frozen embeddings from the linear subspace $\zref$, the nonlinear subspace $\zcont$, and their concatenation $\bm{z}=[\zref;\zcont]$. We apply the same probes to the LeWM embeddings for comparison. Linear probes evaluate whether the information is linearly accessible in the latent space, while nonlinear probes assess whether the information is present but potentially entangled. 
Following the choice of targets and metrics in~\cite{maes2026lewm}, we probe the agent's 2D position in TwoRoom and the 3D positions of the cube and the robot end effector in Cube. We report the mean squared error (MSE) and Pearson's correlation coefficient $r$ between the predictions and the ground truth. 
{As shown in Tabs.~\ref{tab:probe_cube} and ~\ref{tab:probe_tworoom}, $\bm{z}=[\zref;\zcont]$  learned by \ours yields lower mean MSE and higher mean Pearson $r$ than LeWM for agent position in TwoRoom and for 6 of the 8 properties in Cube with each probe type. This suggests that physical quantities can be decoded more accurately from our learned  representation.}

\begin{figure}[t]
    \centering
    \includegraphics[width=0.9\columnwidth]{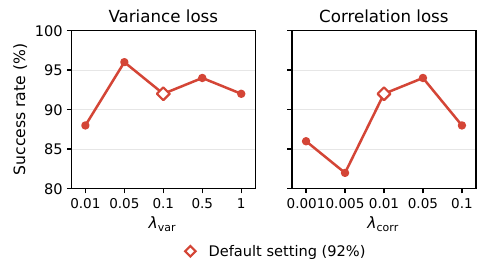}
    \caption{\textbf{Sensitivity to loss weights on Cube.}
    \textbf{Left:} variance weight $\lambda_{\mathrm{var}}$.
    \textbf{Right:} correlation weight $\lambda_{\mathrm{corr}}$.
    Each setting is evaluated on the same 50 short-range planning tasks.
    Open diamonds mark the default setting (92\%).}
    \label{fig:cube-loss-sensitivity}
\end{figure}

\paragraph{Hyperparameter Sensitivity.}
We assess the sensitivity of planning performance to the variance and correlation loss weights by varying one loss weight at a time while keeping all other hyperparameters fixed at their default values. We sweep $\lambda_{\mathrm{var}} \in \{0.01, 0.05, 0.10, 0.5, 1.0\}$ with $\lambda_{\mathrm{corr}} = 0.01$, and $\lambda_{\mathrm{corr}} \in \{0.001, 0.005, 0.01, 0.05, 0.1\}$ with $\lambda_{\mathrm{var}} = 0.10$.
As shown in \cref{fig:cube-loss-sensitivity}, the default weights ($\lambda_{\mathrm{var}} = 0.1$, $\lambda_{\mathrm{corr}} = 0.01$) yield a planning success rate of 92\% , slightly below the highest rate observed in each sweep. \emph{Note that we  use these weights across all four environments without environment-specific tuning.}

{
    \small
    \bibliographystyle{ieeenat_fullname}
    \bibliography{refs}
}

\end{document}